\documentclass[12pt]{article}

\usepackage{amsmath, amssymb}
\usepackage{graphicx}
\usepackage{caption}
\usepackage{subcaption}
\usepackage{multirow}
\usepackage{booktabs}
\usepackage{float}
\usepackage{color}
\usepackage{adjustbox}
\usepackage{enumitem}
\usepackage{algorithm}
\usepackage{algorithmic}
\usepackage{hyperref}
\usepackage{setspace}
\usepackage{lineno}
\usepackage{times}
\usepackage[table]{xcolor}
\usepackage[superscript,biblabel]{cite}

\usepackage{tabularx}

\title{\textbf{ADDT - A Digital Twin Framework for Proactive Safety Validation in Autonomous Driving Systems}}

\author{
    Bo Yu$^{1*}$, Chaoran Yuan$^{2*}$, Zishen Wan$^{3*}$, Jie Tang$^{4\dagger}$, \\
    Fadi Kurdahi$^{2\dagger}$, Shaoshan Liu$^{1\dagger}$\\
    \normalsize{$^{1}$Shenzhen Institute of Artificial Intelligence and Robotics for Society, China}\\
    \normalsize{$^{2}$ University of California, Irvine}\\
    \normalsize{$^{3}$ Georgia Institute of Technology}\\
    \normalsize{$^{4}$South China University of Technology}\\
    \normalsize{$^*$ equal contribution}\\
    \normalsize{$^\dagger$ corresponding author}
}

\date{}

\begin{document}
\sloppy
\title{ADDT - A Digital Twin Framework for Proactive Safety Validation in Autonomous Driving Systems}

\maketitle

% --- ABSTRACT ---
\section*{Abstract}
Autonomous driving systems continue to face safety-critical failures, often triggered by rare and unpredictable corner cases that evade conventional testing. We present the Autonomous Driving Digital Twin (ADDT) framework, a high-fidelity simulation platform designed to proactively identify hidden faults, evaluate real-time performance, and validate safety before deployment. ADDT combines realistic digital models of driving environments, vehicle dynamics, sensor behavior, and fault conditions to enable scalable, scenario-rich stress-testing under diverse and adverse conditions. It supports adaptive exploration of edge cases using reinforcement-driven techniques, uncovering failure modes that physical road testing often misses. By shifting from reactive debugging to proactive simulation-driven validation, ADDT enables a more rigorous and transparent approach to autonomous vehicle safety engineering. To accelerate adoption and facilitate industry-wide safety improvements, the entire ADDT framework has been released as open-source software, providing developers with an accessible and extensible tool for comprehensive safety testing at scale.

% --- TEASER ---
\section*{Teaser}
An autonomous driving design automation platform that identifies rare faults to improve safety, reliability, and real-world readiness.

\section{Introduction}
\label{sec:intro}

Safety remains the foremost challenge facing autonomous driving (AD) systems today. While these systems promise to reduce traffic incidents and revolutionize transportation~\cite{liu2020creating}, the reality is that multiple high-profile accidents have led to fatalities—often triggered by rare and unforeseen “corner cases” that were not captured during the development or validation phases. 

Between 2019 and June 2024, there were 3,979 autonomous vehicle incidents in the United States, resulting in 496 injuries and fatalities. Of these, 83 were fatal. These incidents highlight the critical need for AD systems to effectively handle rare and unpredictable situations~\cite{accidents}. Tesla's Autopilot system has also faced scrutiny. A three-year investigation by the National Highway Traffic Safety Administration (NHTSA) into 956 crashes involving Autopilot found that the system contributed to at least 467 collisions, including 13 fatalities~\cite{accidents2}.

These tragic incidents highlight a fundamental shortcoming in current AD development pipelines: the inability to systematically anticipate, test, and mitigate failure scenarios before deployment. The root of the problem lies in the increasing reliance on end-to-end deep learning architectures. While these models have advanced perception and decision-making capabilities, they operate largely as black boxes~\cite{li2024bevformer, liu2024intelligent}, offering little visibility into how safety-critical decisions are made. As a result, it has become exceedingly difficult for engineers to trace errors, diagnose faults, and validate system behavior under diverse real-world conditions.

This growing gap between system complexity and testing transparency calls for a paradigm shift—from reactive debugging after deployment to proactive, simulation-driven validation during development. To enable this shift, we introduce the \textbf{Autonomous Driving Digital Twin (ADDT)} framework—a high-fidelity, closed-loop simulation platform designed to uncover hidden faults, model rare corner cases, and evaluate safety resilience at scale. By integrating fault injection, real-time performance analysis, and adaptive learning, ADDT provides a new foundation for building trustworthy and robust autonomous driving systems.

Beyond autonomous driving, the ADDT framework extends to the broader field of embodied AI, encompassing service robots, industrial manipulators, agricultural drones, and other intelligent physical systems ~\cite{wang2025eai,wan2025reca}. As these platforms become more complex and safety-critical, ADDT supports scalable, cost-effective development by enabling fault injection, recovery training, and design optimization. Its unified simulation environment facilitates hardware-software co-design while replacing expensive physical testing with rapid, scenario-rich validation. Moreover, its synthetic data generation enhances model robustness across varying environmental conditions, making ADDT a powerful enabler for safe, reliable, and adaptable embodied AI systems.

\subsection{Key Challenges in Autonomous Driving System Development}
\label{subsec:challenges}
As shown in Fig.~\ref{fig:ad_sys}, autonomous driving systems represent one of the most complex engineering endeavors, requiring seamless integration across perception, planning, and control modules, all operating under stringent real-time constraints. Several critical challenges hinder their development:

\begin{figure}[t]
\centering
  \includegraphics[trim=0 0 0 0, clip, width=1.\columnwidth]{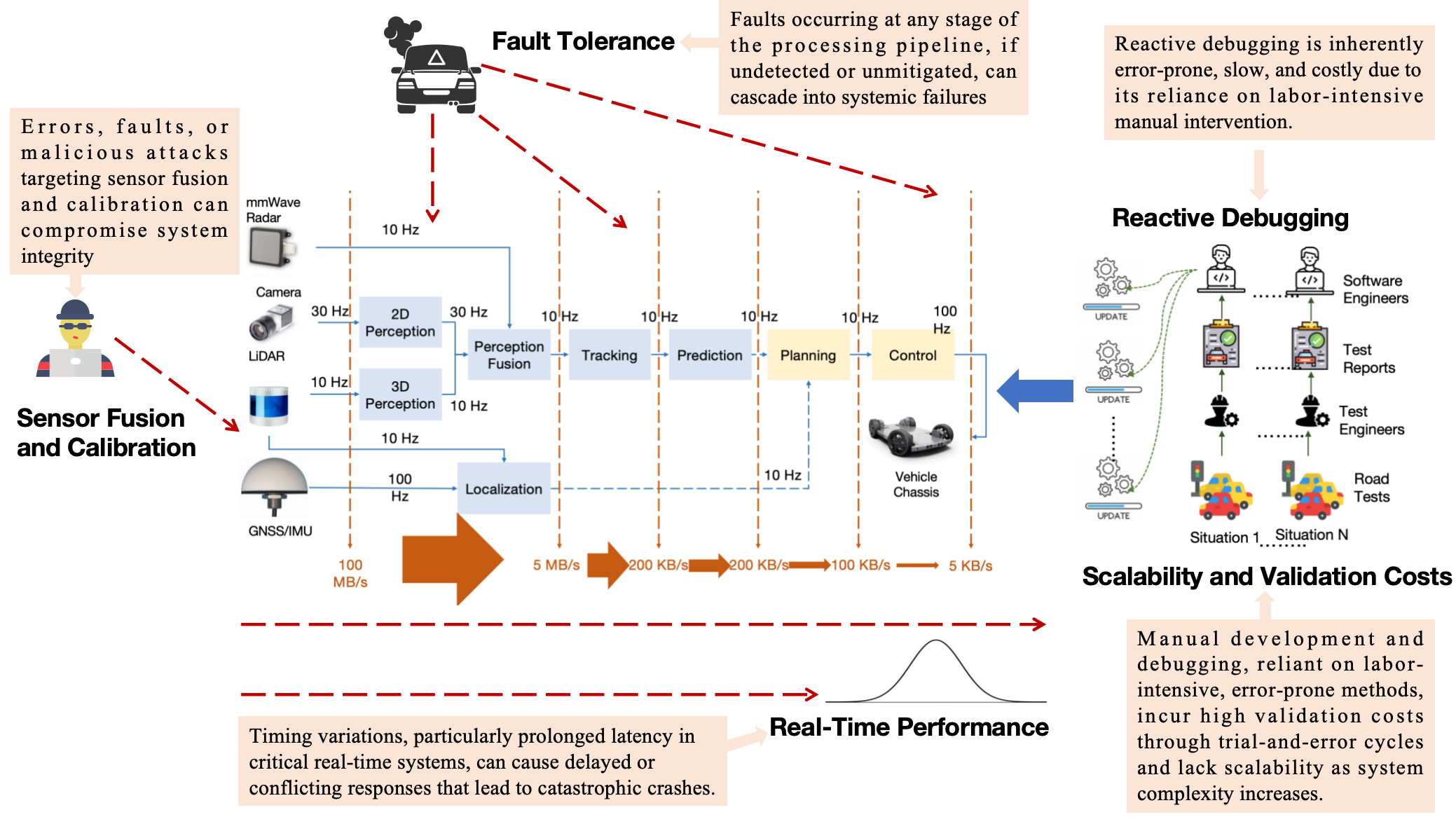}
\caption{Key Challenges in Autonomous Driving System Development.}
\vspace{10pt}
\label{fig:ad_sys}
\end{figure}

\begin{itemize}
    \item \textbf{Fault Tolerance:} Hardware-level faults, such as bit-flips in CPUs or GPUs, and sensor faults, such as data drops and position misalignments, can propagate unpredictably through the software stack, leading to catastrophic failures~\cite{lu2015llfi, tsai2021nvbitfi}.
    \item \textbf{Real-Time Performance:} AD systems must process high-dimensional sensor data streams and make decisions within milliseconds to ensure operational safety~\cite{liu2022rise,chen2025octocache}. Even minor latency spikes can result in safety-critical failures during high-speed maneuvers.
    \item \textbf{Sensor Fusion and Calibration:} Effective sensor fusion, involving the integration of data from diverse sensors (e.g., cameras, LiDAR, radar), is crucial for the safety of autonomous driving systems, as it ensures accurate and reliable environmental perception. Challenges such as temporal misalignment, calibration drift, and sensor noise can significantly degrade the quality of the combined sensor data, potentially leading to incorrect interpretations of the driving environment, impaired decision-making, and ultimately increased risk of accidents ~\cite{donner2009empirical}.    
    \item \textbf{Scalability and Validation Costs:} Physical testing is expensive, time-consuming, and limited in its ability to replicate diverse environmental scenarios or rare fault conditions~\cite{wu2023autonomy}.
    \item \textbf{Reactive Debugging Paradigm:} Current AD system validation largely relies on post-deployment debugging rather than proactive fault modeling during the design phase~\cite{wan2021survey}. Recent approaches emphasize integrating formal verification techniques into AI-driven systems to systematically address these shortcomings~\cite{seshia2022toward}.
\end{itemize}

These challenges underscore the need for systematic, scalable tools capable of proactively addressing performance, reliability, and fault resilience during the design and validation phases of AD system development.

\subsection{The Digital Twin-Based Design Automation Paradigm}

The challenges outlined in subsection~\ref{subsec:challenges} are not unique to autonomous driving. Similar barriers—complexity, cost, and scalability—were encountered during the early stages of integrated circuit development. The turning point in that domain was the advent of Electronic Design Automation (EDA), which introduced automated workflows, fault modeling, and scalable simulation infrastructure~\cite{history}. This transformation demonstrates that when system complexity surpasses the limits of manual engineering, automation and simulation become indispensable tools for productivity and reliability.

Today, the autonomous driving (AD) industry finds itself at a comparable inflection point. Despite significant algorithmic advances, AD system development remains heavily reliant on physical road testing and manual debugging~\cite{wu2023autonomy}. To enable the next leap in productivity and safety, a transition toward simulation-driven, design automation methodologies is essential. This motivates the introduction of the \textbf{Digital Twin-Based Design Automation Paradigm}.

At the heart of this paradigm lies the concept of a \textbf{digital twin}—a high-fidelity virtual representation of a physical AD system that enables repeatable, controllable experimentation across a wide array of real-world conditions~\cite{aistrategy2023}. These digital environments not only reduce costs and risks but also expose systems to rare and safety-critical scenarios that are difficult to replicate in physical testing.

To operationalize this paradigm, we present the \textbf{Autonomous Driving Digital Twin (ADDT)} framework, which incorporates the following key components:

\begin{itemize}
    \item \textbf{3D Digital Twin Mapping:} The ADDT uses high-resolution 3D reconstructions of urban and suburban environments, built upon HD maps and OpenDRIVE standards~\cite{yu2022digitaltwin}. These maps capture fine-grained details such as lane markings, traffic signals, road curvature, and elevation, allowing for realistic and geographically accurate simulation environments that reflect the challenges encountered in real-world driving.

    \item \textbf{Traffic Control Modules:} Built upon the OpenSCENARIO specification~\cite{openscenario}, these modules support parameterized modeling of traffic participants including cars, cyclists, and pedestrians, with programmable behaviors such as lane changes, sudden stops, and cut-ins. The ability to synthesize both nominal and adversarial traffic flows enables robust validation of planning and control algorithms under varied and dynamic conditions.

    \item \textbf{Sensor Modeling Modules:} These modules emulate the behavior of real-world sensors such as LiDAR, cameras, radar, and ultrasonic detectors. They incorporate realistic noise distributions, temporal synchronization issues, field-of-view limitations, and calibration drift~\cite{donner2009empirical}. This allows for accurate perception testing and analysis of how sensor-level imperfections impact downstream modules in the autonomy stack.

    \item \textbf{Vehicle Dynamics Models:} Using physics-based simulation engines such as CarMaker~\cite{carmaker}, ADDT simulates vehicle motion under realistic mechanical constraints, including suspension models, tire forces, and actuator delays. These dynamics models ensure that control outputs generated by the AD system are tested in physically plausible and safety-critical scenarios, including high-speed cornering and emergency braking.

    \item \textbf{Fault Injection Mechanisms:} ADDT introduces targeted and repeatable faults at both hardware and software levels, such as CPU/GPU bit-flip errors, memory corruption, sensor data drops, and misalignments~\cite{lu2015llfi, tsai2021nvbitfi}. These mechanisms enable rigorous fault resilience testing, offering insights into system failure modes and allowing developers to evaluate safety margins and recovery capabilities.
\end{itemize}

%In addition to deterministic scenario modeling, ADDT incorporates \textbf{reinforcement learning (RL)} techniques to explore state-action spaces and identify safety-critical edge cases~\cite{feng2023dense}. 

These methods allow the AD system to adaptively refine safety policies in the presence of uncertainties, such as unpredictable agent behaviors or sensor degradation, thus expanding the scope of validation beyond static test cases.

Altogether, the ADDT framework and the Digital Twin-Based Design Automation Paradigm together represent a scalable, proactive, and cost-effective foundation for building safer and more reliable autonomous driving systems.

\subsection{Contributions}

This work introduces the \textbf{Autonomous Driving Digital Twin (ADDT)} framework as a transformative simulation-based design automation paradigm for autonomous driving systems. Inspired by the historical impact of Electronic Design Automation (EDA), ADDT systematically addresses critical development challenges in performance evaluation, fault resilience, and cost-effective validation. The major contributions are as follows:

\begin{itemize}
    \item \textbf{Design Automation Paradigm:} We propose a digital twin-based design automation framework that enables proactive, closed-loop simulation of AD systems, shifting the development process from reactive, physical testing to scalable, virtual-first validation.
    
    \item \textbf{Comprehensive Fault Injection and Analysis:} ADDT supports systematic fault injection at both sensor and compute levels, offering fine-grained, scenario-specific evaluation of how hardware and perception faults affect mission performance, planning accuracy, and control robustness.
    
    \item \textbf{Scalable and Cost-Effective Validation:} The framework achieves orders-of-magnitude improvements in testing speed and cost-efficiency compared to physical testing, enabling large-scale scenario evaluation and safety validation at a fraction of the cost.
\end{itemize}

%In addition, ADDT integrates reinforcement learning-based safety validation strategies~\cite{feng2023dense}, further enhancing its ability to model system adaptation under dynamic and uncertain conditions. Together, 
These capabilities establish ADDT as a foundational tool for advancing safe, reliable, and efficient autonomous driving system development.
Specifically, to foster community collaboration and accelerate industry adoption, the ADDT framework has been fully open-sourced. This ensures that automotive OEMs and Tier-1 suppliers can readily access, extend, and integrate the tool into their own development pipelines for testing, validation, and safety assurance of autonomous driving products.

\section{Results}
\label{sec:res}

In this section, we present the quantitative findings of the Autonomous Driving Digital Twin (ADDT) framework, focusing on latency variations, sensor-level reliability, and compute-level reliability across diverse driving scenarios. The results highlight the ADDT's capability to systematically evaluate critical design challenges in autonomous driving (AD) systems and offer insights into addressing them proactively.

\subsection{Testing scenarios}
\begin{figure}[t]
    \centering
    % Subfigure 0 (Parameters)
    \begin{subfigure}{0.88\columnwidth}
        \captionsetup{labelformat=empty, font=scriptsize, aboveskip=-8pt, belowskip=2pt}
        \includegraphics[width=\columnwidth]{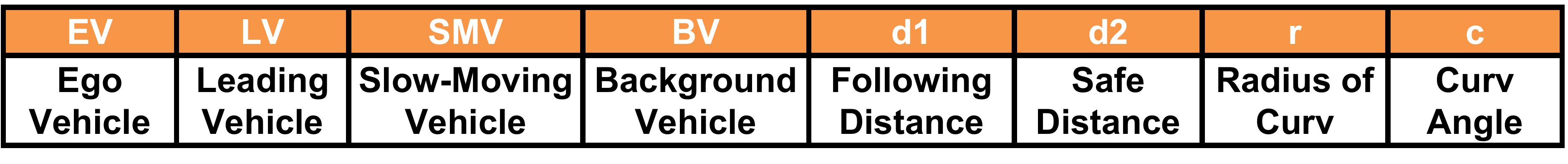}
        \vspace{-9pt}
        %\caption{Parameter settings for evaluation scenarios.}
        \label{fig:Parameters}
    \end{subfigure}
    \vspace{0pt} % Adjust spacing between rows

    % Subfigure 1
    \begin{subfigure}{0.325\columnwidth}
        \captionsetup{labelformat=empty, font=scriptsize}
        \includegraphics[width=\columnwidth]{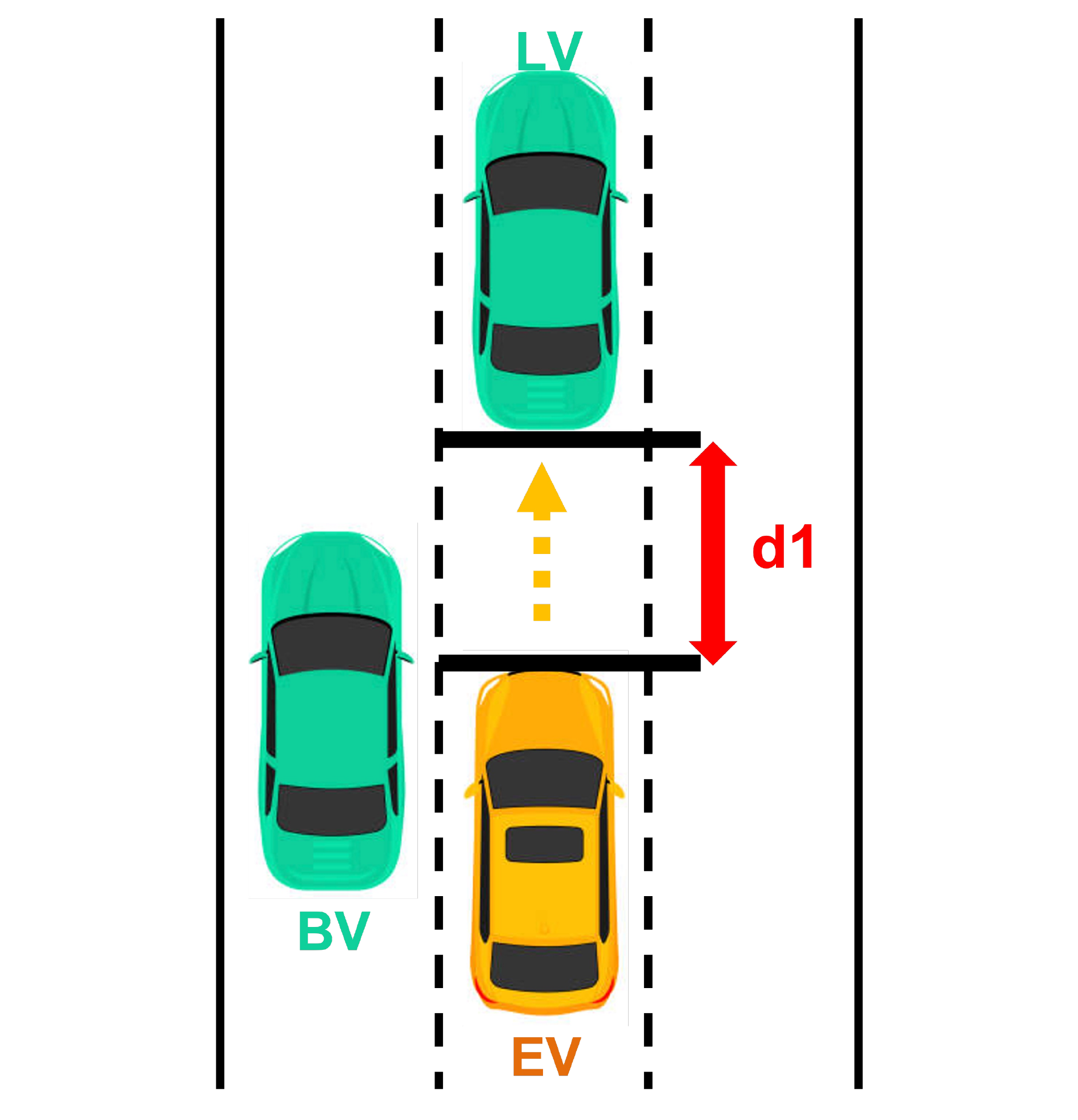}
        \caption{(1) Vehicle following.}
        \label{fig:Scene1}
    \end{subfigure}
    \hfill
    % Subfigure 2
    \begin{subfigure}{0.325\columnwidth}
        \captionsetup{labelformat=empty, font=scriptsize}
        \includegraphics[width=\columnwidth]{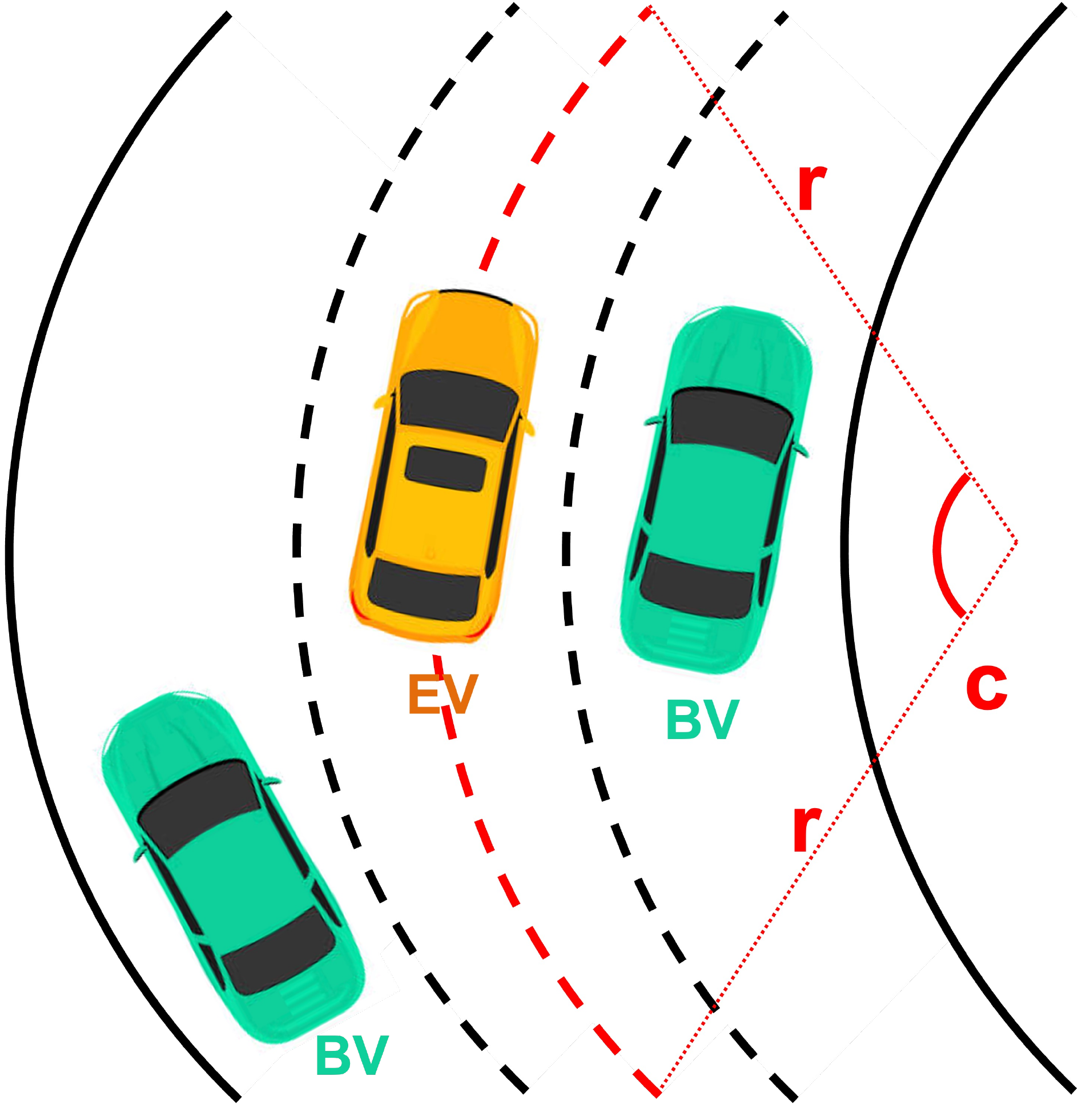}
        \caption{(2) Large-curvature turning.}
        \label{fig:Scene2}
    \end{subfigure}
    \hfill
    % Subfigure 3
    \begin{subfigure}{0.325\columnwidth}
        \captionsetup{labelformat=empty, font=scriptsize}
        \includegraphics[width=\columnwidth]{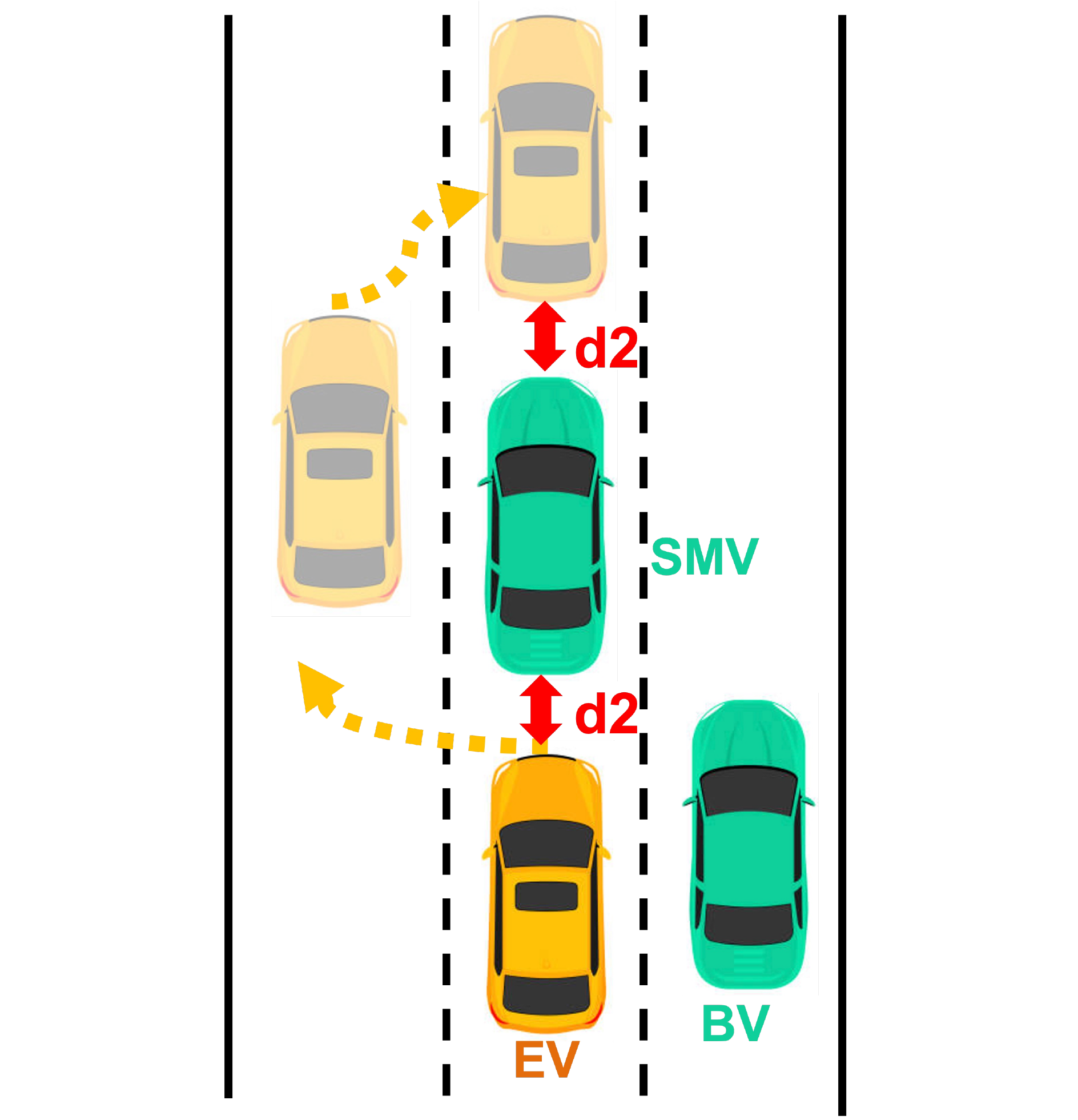}
        \caption{(c) Overtaking.}
        \label{fig:Scene3}
    \end{subfigure}

    \caption{Driving scenarios for evaluation.}
    %(EV: Ego Vehicle, BV: Background Vehicle).
    \label{fig:Scenarios}
\end{figure}

To rigorously assess the safety of AD computing systems, we define three categories of driving scenarios that encompass most daily highway driving situations: 1) vehicle following, with or without emergency braking of the leading vehicle, 2) large-curvature turning, and 3) overtaking. Fig.~\ref{fig:Scenarios} illustrates these three scenario types and their associated parameters, such as inter-vehicle distances and vehicle speeds. By parameterizing and varying these parameters, the ADDT can generate a large-scale synthetic dataset representing diverse driving conditions. In addition to synthetic data, we utilize the real-world nuScenes dataset~\cite{nuScenes} to compare the effectiveness of our ADDT environment. It's important to note that because real-world datasets are static captures of driving scenarios and cannot facilitate real-time closed-loop control, we use them solely to evaluate computing latency variations in our experiments.

\subsection{Safety Related to Latency Variation}
AD systems are subject to stringent real-time computing constraints to ensure timely vehicle responses to environmental changes. The computing latency of an AD system, from the detection of a safety-critical event by a sensor to the execution of a response action, should be less than 100 milliseconds to achieve human-level safety~\cite{luo2019time}. 

\begin{figure*}[t]
    \centering
    % Subfigure 1
    \begin{subfigure}{0.48\textwidth}
        \captionsetup{labelformat=empty, font=small}
        \includegraphics[width=\textwidth]{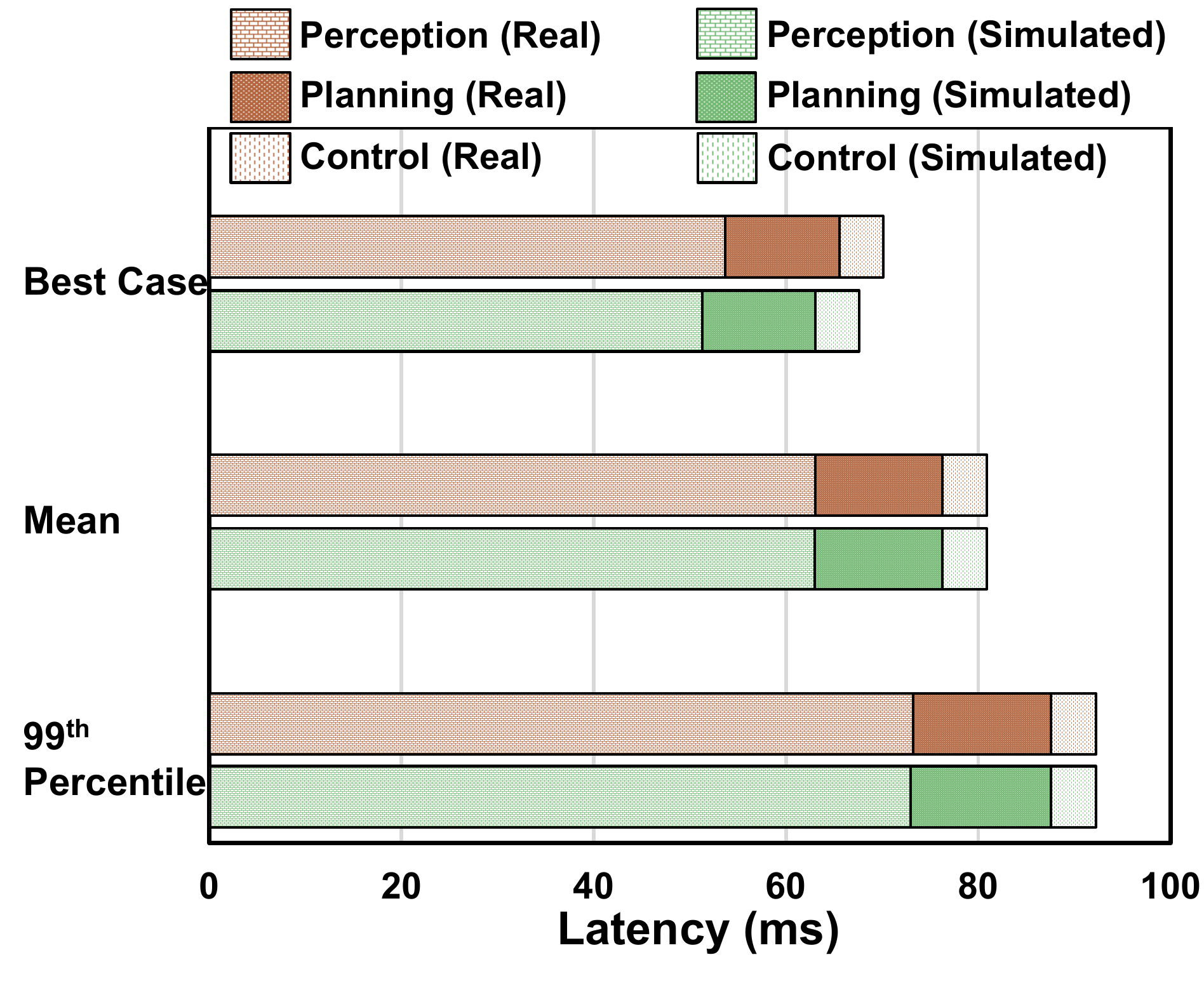}
        \caption{(a) End-to-End Latency Comparison.}
        \label{fig:Latency_3cases}
    \end{subfigure}
    \hfill
    % Subfigure 2
    \begin{subfigure}{0.5\textwidth}
        \captionsetup{labelformat=empty, font=small}
        \includegraphics[width=\textwidth]{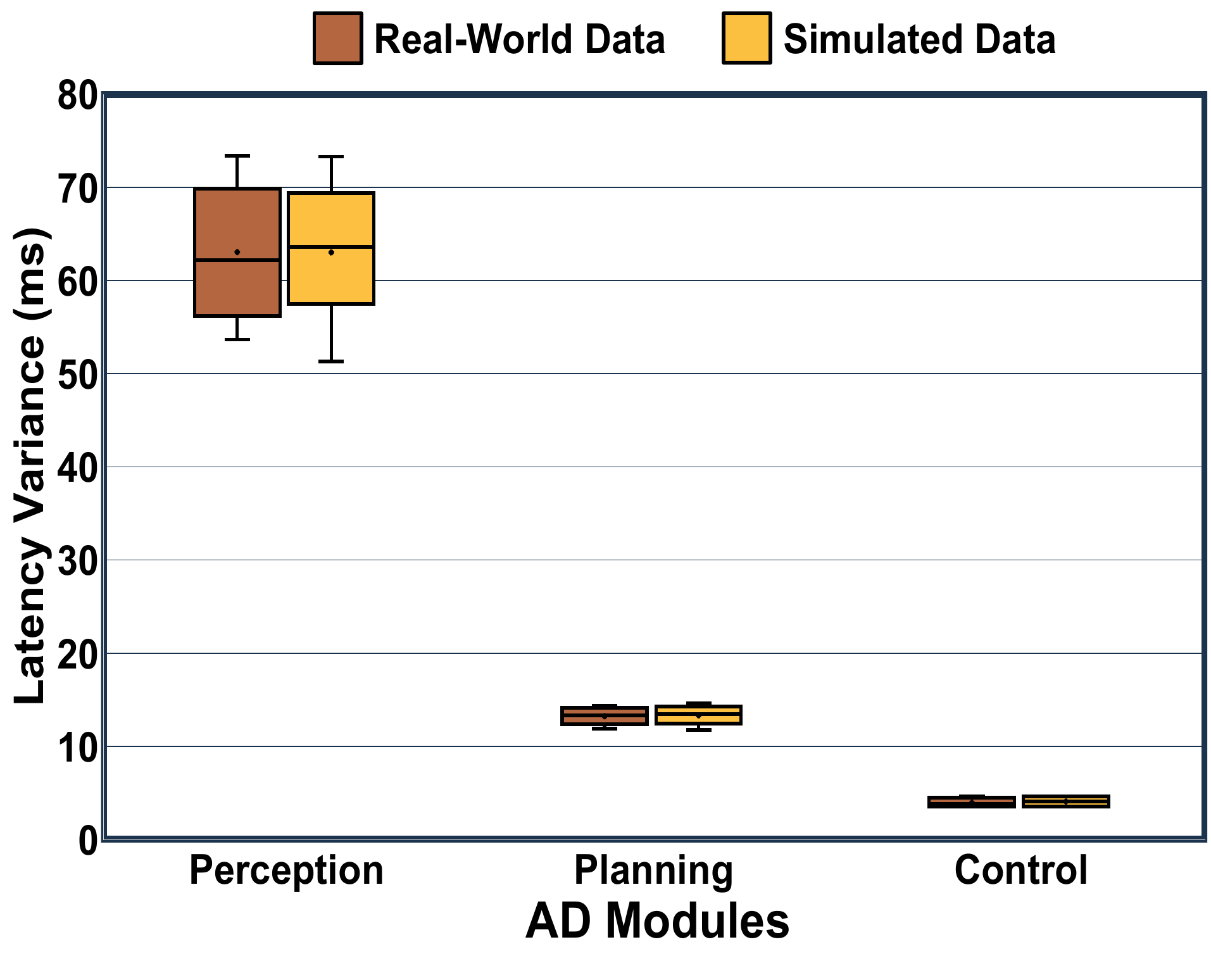}
        \caption{(b) Latency Distribution of AD Modules.}
        \label{fig:Latency_modules}
    \end{subfigure}
    \hfill
    \caption{Latency Distribution Comparison between Real-World and Simulated Data.}
    %\vspace{14pt}
    \label{fig:variance}
\end{figure*}

\textbf{Latency Distribution.} 
A significant safety concern in AD systems is the occurrence of large variations in computing latency~\cite{yu2020building}. To investigate this latency variation using ADDT, we benchmarked latency measurements from ADDT simulations against real-world datasets. Fig.~\ref{fig:Latency_3cases} compares the best, average, and 99th percentile end-to-end latency of the AD software between simulated and real-world data. We also conducted a statistical analysis of the latency distribution for each AD module, as shown in Fig.~\ref{fig:Latency_modules}. The latency distributions generated by ADDT closely match those observed in real-world data, with a KL divergence of 0.1, thereby validating the simulation environment's fidelity.

\textbf{Impact of Workload on Latency Variation.} 
Latency variations in AD systems arise primarily from dynamic workloads influenced by environmental complexity and computational resource availability~\cite{gan2021eudoxus,gong2024lotus}. Ensuring predictable and bounded latency is essential for real-time decision-making and operational safety~\cite{yu2020building}.

The number of traffic objects significantly influences the workloads of AD modules~\cite{gong2024lotus,gan2021eudoxus}. However, real-world datasets often have limitations in the number of detectable objects per scene, hindering the conclusive identification of conditions that violate real-time constraints.

\begin{figure}[H]
    \centering
    \includegraphics[width=0.8\textwidth]{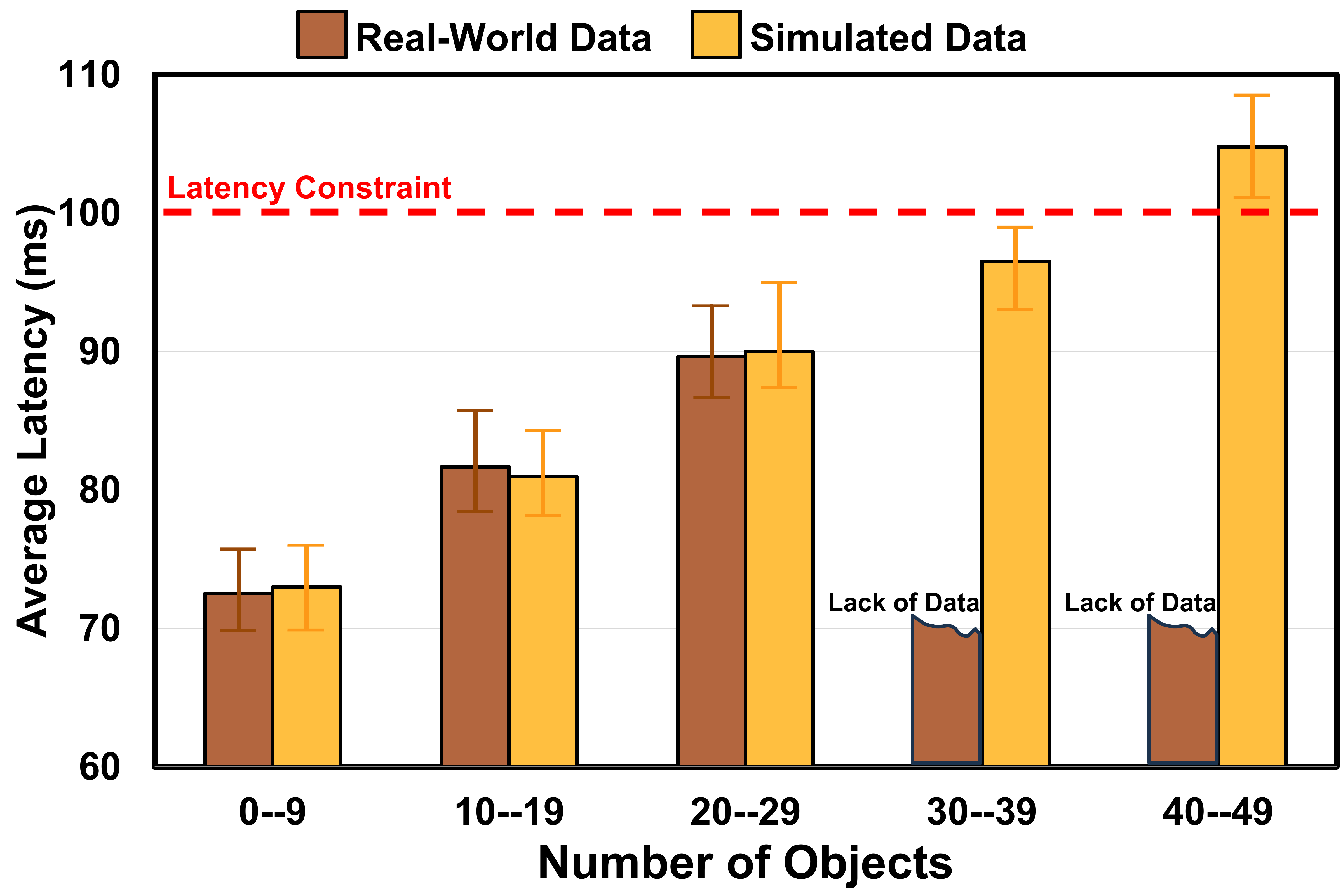}
    \caption{Latency variation under different object counts in driving scenarios.}
    \label{fig:latency_workload}
\end{figure}

ADDT addresses this limitation by synthesizing a controllable number of objects, allowing for the extrapolation of studies on how the number of detectable objects impacts computing latency. Fig.~\ref{fig:latency_workload} illustrates latency performance under varying object counts. When the object count exceeded 40 per frame, latency consistently surpassed the critical 100 ms threshold, compromising real-time performance. These observations align with recent findings in latency optimization frameworks for edge devices, where adaptive workload management strategies are employed to minimize latency variations under dynamic processing demands~\cite{gong2024lotus}.

\subsection{Sensor Level Faults Simulation}
The sensing subsystem in an AD system must provide temporally and spatially aligned multi-sensor data streams~\cite{liu2021brief}. Sensor faults in either the temporal or spatial domains can degrade the quality of these data streams, leading to safety issues. In the temporal domain, sensor faults manifest as frame rate jitters or frame drops, often caused by hardware malfunctions such as camera breakdowns~\cite{PETR}. These faults can result in missed object detections and potential safety hazards. In the spatial domain, sensor orientation shifts, often caused by external forces during driving~\cite{AVFI}, increase spatial misalignment errors, degrading object position estimation accuracy within the AD system.
%sensor faults, typically caused by shifts in sensor orientation due to external forces during prolonged driving~\cite{AVFI}, can increase errors in the spatial alignment of sensor data. This, in turn, degrades the accuracy of object position estimation within the AD system.

Using ADDT, we can efficiently inject sensor faults, such as data drops and sensor shifts, to assess their impact on perception module performance and overall AD system safety.

\textbf{Sensor Data Drops.}  
Table~\ref{table:frame_dropping} illustrates the relationship between mission success rates and camera frame drop rates. In our experiment, AD vehicles driving off-lane or crashing with other vehicles are considered mission failures. Our experimental results demonstrate a strong correlation between sensor integrity and AD safety. Notably, at the same frame drop rate, safety performance varies across different scenarios. Overtaking scenarios, for example, exhibit higher sensitivity to frame drops due to their dynamic nature and reliance on timely reactions. 

\begin{table}[t]
\centering
\caption{Driving Task Success Rates under Different Frame Drop Rates}
\scriptsize
\setlength{\tabcolsep}{2pt}
\begin{tabularx}{\columnwidth}{|>{\centering\arraybackslash}X|>{\centering\arraybackslash}X|>{\centering\arraybackslash}X|>{\centering\arraybackslash}X|>{\centering\arraybackslash}X|}
\toprule
\textbf{Frame drop rate} & \textbf{1\%} & \textbf{2\%} & \textbf{5\%} & \textbf{10\%} \\ 
\midrule
Large-curv. turning & 100\% & 99.58\% & 96.25\% & 88.75\% \\ 
Vehicle following & 99.58\% & 99.58\% & 97.92\% & 94.16\% \\ 
Overtaking & 97.08\% & 87.50\% & 73.75\% & 54.58\% \\ 
\bottomrule
\end{tabularx}
\label{table:frame_dropping}
\end{table}

\begin{figure}[t]
    \centering
    % Subfigure 1
    \begin{subfigure}{0.48\columnwidth}
        \captionsetup{labelformat=empty, font=small}
        \includegraphics[width=\columnwidth]{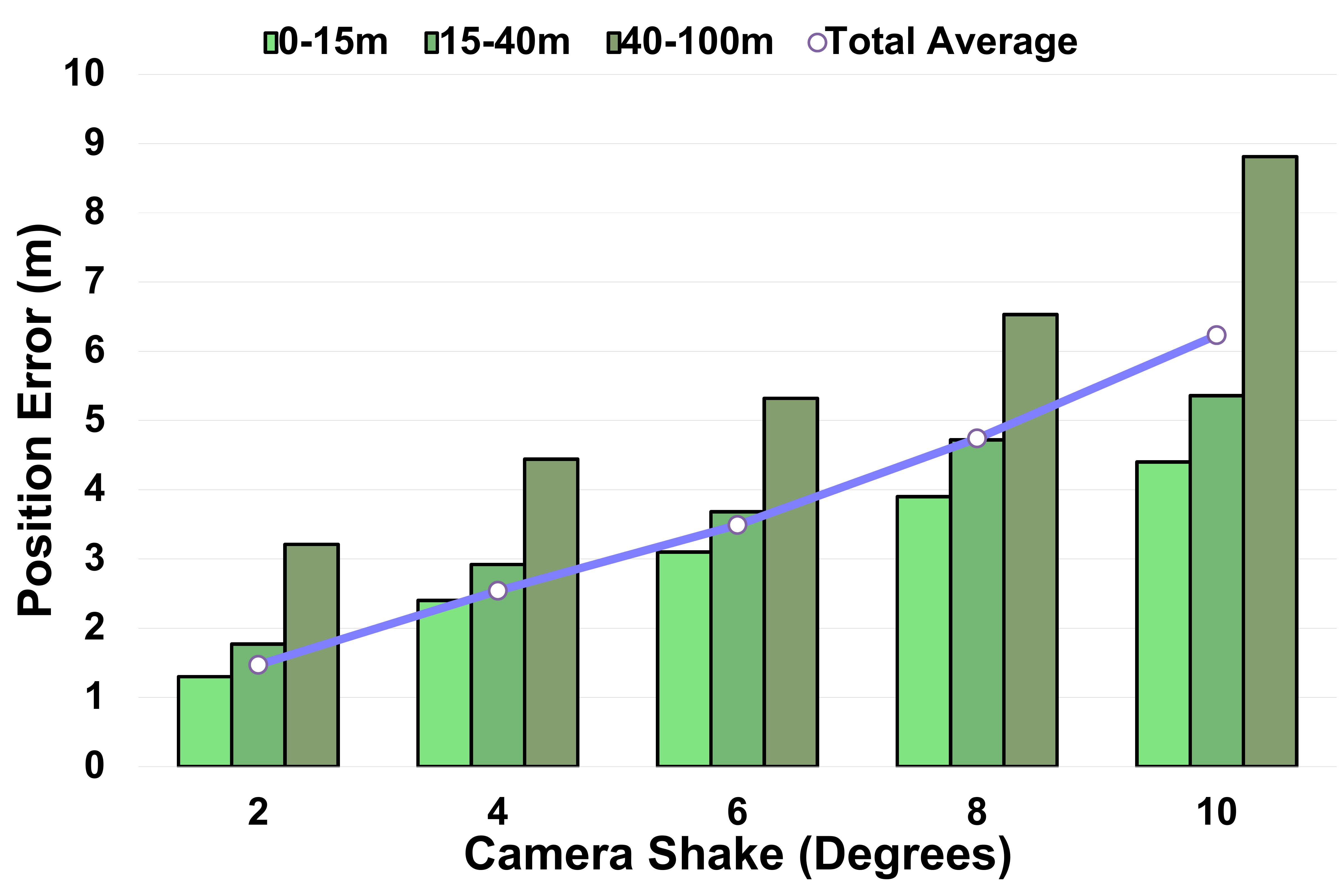}
        \caption{(a) Average positional error of detected objects.}
        \label{fig:Scene1}
        %\vspace{14pt}
    \end{subfigure}
    \hfill
    % Subfigure 2
    \begin{subfigure}{0.5\columnwidth}
        \captionsetup{labelformat=empty, font=small}
        \includegraphics[width=\columnwidth]{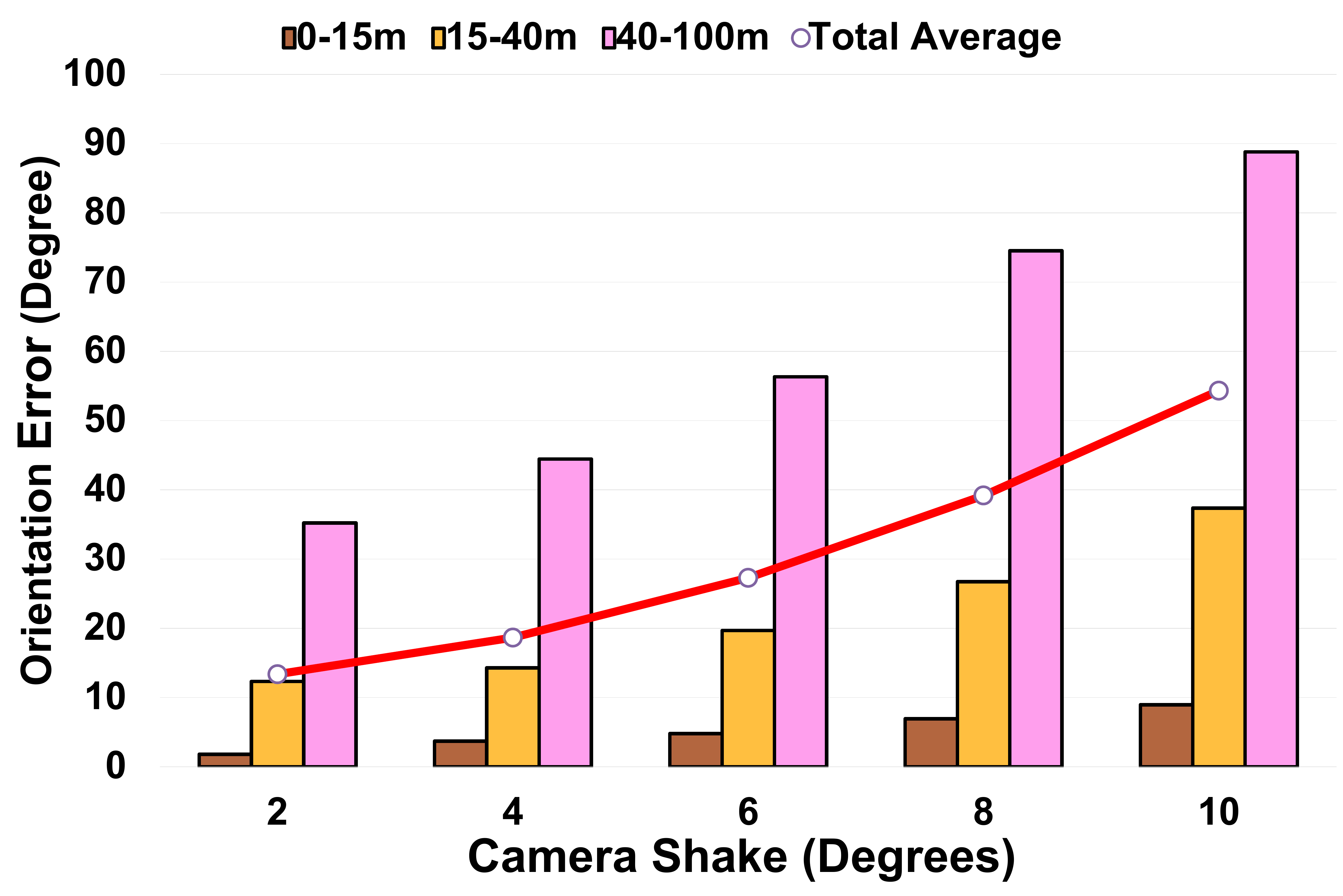}
        \caption{(b) Average orientational errors of detected objects.}
        \label{fig:Scene2}
    \end{subfigure}
    \hfill
    %\vspace{8pt}
    \caption{The relationship between camera orientation shift and the positional and orientational errors of detected objects.}
    %\vspace{8pt}
    \label{fig:camera_shake}
\end{figure}

\textbf{Sensor Shifts.}
Fig.~\ref{fig:camera_shake} illustrates the relationship between camera orientation shifts and the accuracy of the perception algorithm under various sensor shifts. For each camera shake level, we conducted 100 independent experiments (i.e., $N = 100$) to compute the average positional and orientational errors. The average position error ($E_p$) is the mean Euclidean distance between the detected object's center $(x_d, y_d)$ and the ground truth center $(x_{gt}, y_{gt})$ over $N$ experiments:
\begin{equation}
    E_p = \frac{1}{N} \sum_{i=1}^{N} \sqrt{(x_{d}^{(i)} - x_{gt}^{(i)})^2 + (y_{d}^{(i)} - y_{gt}^{(i)})^2}
    \label{eq:position_error}
\end{equation}
Similarly, the average orientational error ($E_\theta$) is the mean absolute difference between the detected object's orientation ($\theta_d$) and the ground truth orientation ($\theta_{gt}$):
\begin{equation}
    E_\theta = \frac{1}{N} \sum_{i=1}^{N} |\theta_{d}^{(i)} - \theta_{gt}^{(i)}|
    \label{eq:orientation_error}
\end{equation}

The experimental results show that as camera orientation errors increase, the position and orientation errors of detected objects also increase. Furthermore, the distance of objects impacts perception accuracy; at a given level of camera shift error, more distant objects exhibit greater perception errors. These findings highlight the importance of robust sensor calibration mechanisms and fault-tolerant perception algorithms in mitigating the impact of sensor errors~\cite{donner2009empirical, FastBEV}.

\subsection{Compute Level Faults Simulation}
Compute-level faults in this study are defined as bit-flip errors occurring in the memory, CPUs, and GPUs of the computing system. These subtle hardware errors are difficult to identify and diagnose in AD systems because AD algorithms, particularly deep neural networks, exhibit a degree of fault tolerance to them~\cite{wan2023berry}. Paradoxically, this inherent fault tolerance makes bit-flip errors a significant potential hazard to AD system safety, as they may not consistently propagate to affect the AD system's output.
 
\begin{figure}[H]
    \centering
    \includegraphics[width=1.0\textwidth]{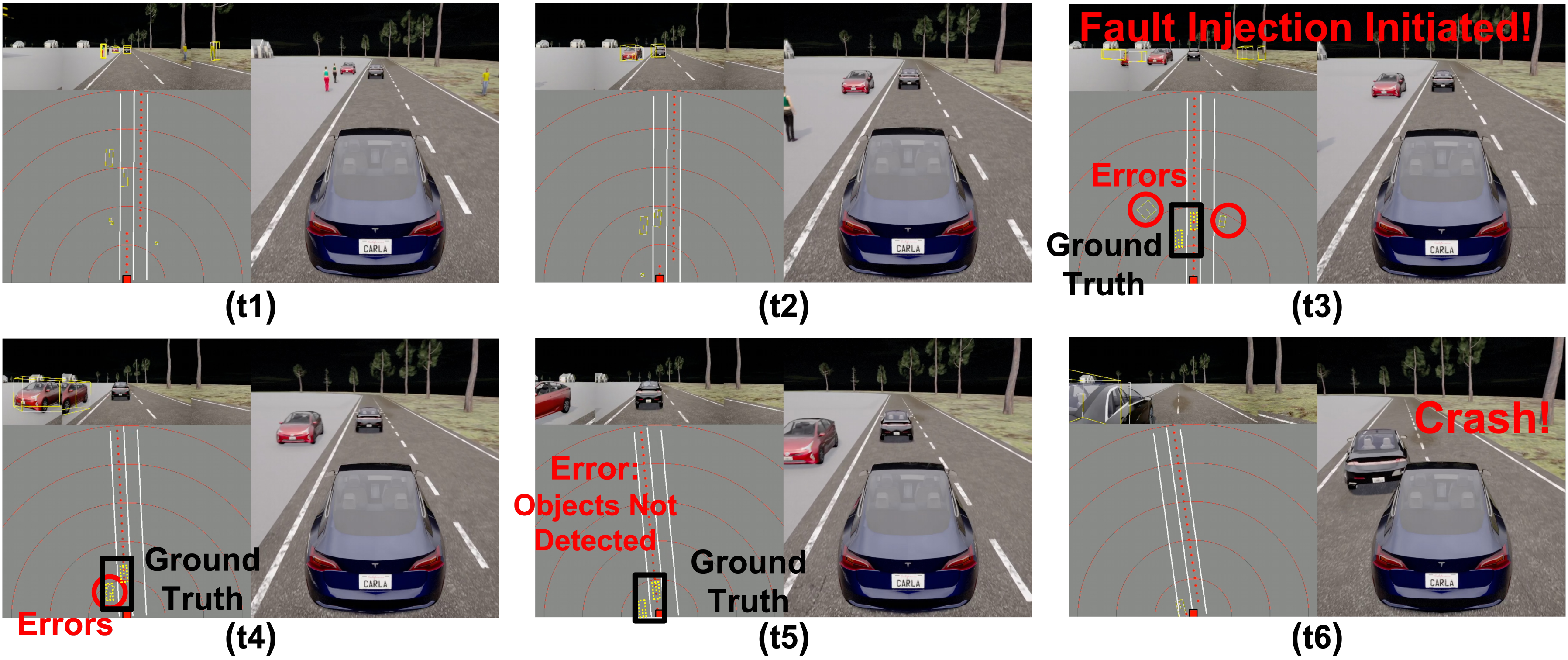}
    %\caption{Demonstration of compute-level fault injection process: the ego vehicle performed a lane-changing maneuver to overtake a slow-moving vehicle, while compute-level bit-flip faults were injected into the BEV perception module between frames (t2) and (t3), leading to errors in object detection and localization. Notably, the slow-moving vehicle was incorrectly undetected in multiple frames, such as (t3), (t4) and (t5), causing false free-space estimation. As a result, the planning module generated incorrect decisions, culminating in a crash at frame (t6).}
    \caption{Demonstration of compute-level fault injection in an overtaking scenario: Between frames t2 and t3, bit-flip faults were injected into the perception module, causing errors in object detection and localization. The vehicle ahead of the autonomous vehicle was undetected in frames t3, t4, and t5, leading to a false free-space estimation. Consequently, the planning module made incorrect decisions, resulting in a crash at frame t6.}
    \label{fig:Fault_Demo}
\end{figure}

Analyzing and studying bit-flip errors requires a large volume of experiments to statistically evaluate their impact on safety, making physical testing extremely time-consuming~\cite{hsiao2023silent,hsiao2023mavfi}. ADDT, by combining digital twin-based real-time Hardware-in-the-Loop (HIL) simulation with hardware fault injection capabilities, provides the means to simulate various bit-flip errors within the AD system and analyze their safety impacts across diverse driving scenarios. This facilitates the controllable, efficient, and safe evaluation of their effects on system safety. Fig.~\ref{fig:Fault_Demo} illustrates a fault injection process into the perception module within the ADDT environment, where bit-flip errors result in incorrect perception and a subsequent crash.

\begin{figure}[H]
    \centering
    \includegraphics[width=1.\textwidth]{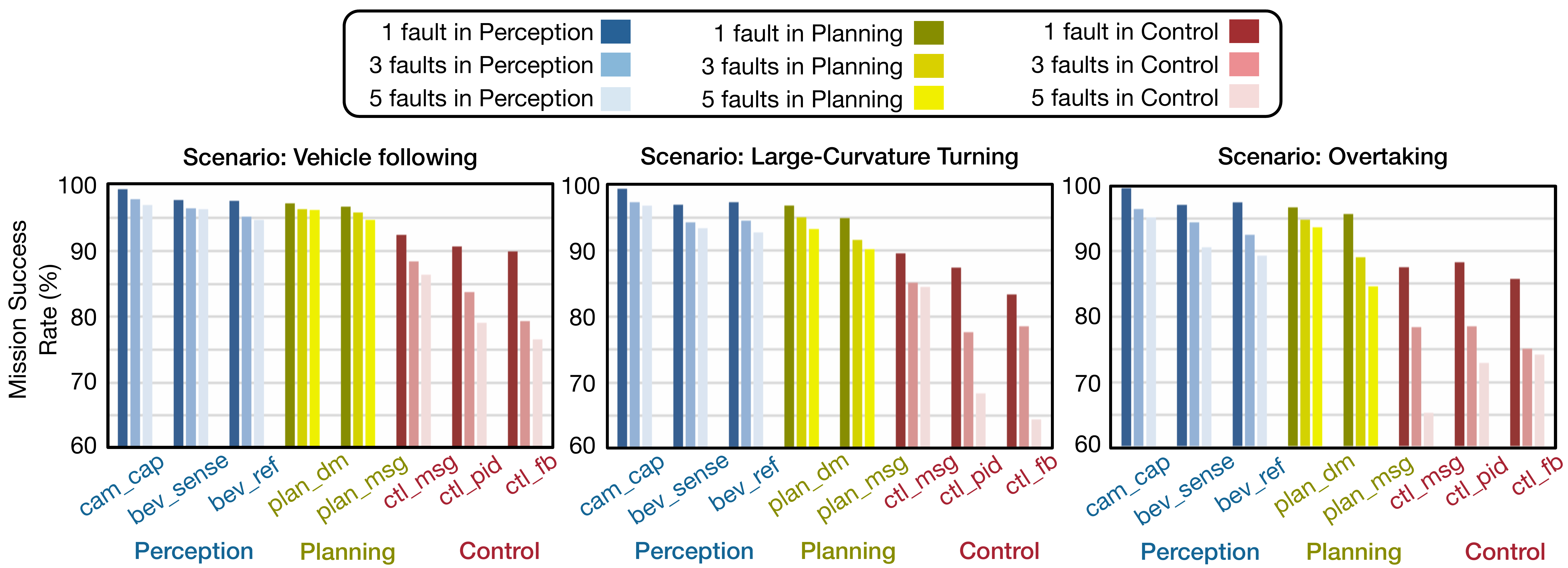}
    \caption{
    Relationship between mission success rates and the number of faults in different driving scenarios and AD software nodes.}
    \label{fig:bit_flip_results}
\end{figure}

We conducted a thorough evaluation of bit-flip error safety impacts, examining the effects of error quantity and location. Varying numbers of bit-flip errors were injected into ROS nodes (see node definition in Fig.~\ref{fig:graph_nodes}) across modules within the AD pipeline for each driving scenario, with results illustrated in Fig.~\ref{fig:bit_flip_results}. Key findings are summarized below:

\begin{enumerate}

\item \textbf{Module-Level Reliability Variations:}
The AD software stack exhibits inherent reliability variations across modules. In particular, control nodes are the most sensitive to bit-flips, experiencing the most significant drops in mission success rates. For example, in the large-curvature turning scenario, success drops from 90.67\% with one control fault to just 66.90\% with five faults. This underscores the criticality of the control stage, which is responsible for real-time actuation and trajectory execution, where even minor errors can destabilize the vehicle -- especially in scenarios requiring precise maneuvering.

% The results reveal that the AD pipeline exhibits inherent reliability variations - control nodes are particularly vulnerable to hardware bit-flip faults, experiencing the most significant drops in mission success rates. This sensitivity is especially pronounced in large curvature turning, where mission success rates decline sharply from 90.67\% with one fault to just 66.90\% with five faults. This suggests that the control stage of the AD pipeline, responsible for executing driving decisions in real time, is highly susceptible to hardware-induced errors. Such errors can severely compromise the vehicle’s ability to maintain stability and trajectory, particularly in complex driving conditions that demand precise control.

\item \textbf{Fault Propagation and Pipeline Sensitivity:}
Moreover, the data underscores the fact that faults occurring closer to the end of the AD pipeline tend to have a disproportionately larger impact on mission success. This is likely because errors introduced earlier in the pipeline can often be masked or mitigated by subsequent processing stages. In contrast, faults in the later stages (particularly in control modules) directly affect vehicle actuation, where even minor inaccuracies are less likely to be corrected downstream. These findings highlight the urgent need for fault-tolerant design and runtime validation mechanisms in control software to ensure system robustness under hardware-level perturbations.

% direct involvement in the final execution of driving commands, where even minor discrepancies can lead to significant performance degradation. The amplified effect of these faults highlights the critical importance of implementing robust fault tolerance and error-checking mechanisms within the control systems to mitigate their impact. 

\item \textbf{Scenario-Level Reliability Variations:}
The sensitivity to faults is not uniform across all driving scenarios, further emphasizing the need for scenario-specific testing and robustness strategies. For instance, in the lane-changing scenario, planning nodes exhibit higher vulnerability to faults compared to other scenarios, with mission success rates falling notably as the number of faults increases. This indicates that the planning stage, which involves determining the vehicle’s path and making decisions on lane changes, is critical in scenarios where quick and accurate decision-making is required. On the other hand, in large curvature turning, the control nodes are more significantly impacted, possibly due to the increased demand for precise vehicle maneuvering in such conditions.
\end{enumerate}

\subsection{Cost and Efficiency Comparison}
The ADDT framework offers significant advantages over traditional physical road testing by reducing both hardware costs and testing timelines.

Our ADDT system typically costs less than \$5,000 and includes a workstation (configured with an Intel i7-9800X CPU and an NVIDIA GTX 2080 Ti GPU) running the digital twin environment, an embedded computing platform (an Nvidia Jetson Xavier) for the AD software, and the necessary network connections. In contrast, an AD testing vehicle, such as a Tesla Model Y, typically costs more than \$50,000. Therefore, ADDT is approximately 10 times more cost-effective than using physical AD vehicles.

Furthermore, ADDT significantly enhances testing efficiency for scenario-based safety analysis. Firstly, simulation tests can run continuously, 24 hours a day, while physical road testing, dependent on human testers, is limited to typical 8-12 hour workdays. Secondly, ADDT's lower hardware cost enables more tests within the same budget. Thirdly, setting up test scenarios is considerably more efficient in ADDT than the time-consuming process of real-world scenario preparation. Assuming real-world scenario setup takes 3 times longer than the testing process itself, and that ADDT allows for 10 times the number of tests for the same cost, while physical testing is limited to an 8-hour workday, ADDT is approximately 90 times more time-efficient than physical testing, given equivalent testing costs and time constraints.

\section{Discussion}
\label{sec:dis}

The ADDT framework introduces two critical innovations to address persistent challenges in autonomous driving development. First, its comprehensive digital twin infrastructure proactively identifies safety-critical issues such as latency violations, sensor misalignments, and compute-level faults—corner cases frequently missed by existing methods. Second, ADDT's scalable simulation-first approach significantly reduces reliance on costly and limited-scope physical testing, greatly enhancing development efficiency and affordability. This combination establishes ADDT as a robust solution for improving safety and reliability in autonomous driving systems. Below, we elaborate on these contributions, compare ADDT to existing industry tools, identify current limitations, and outline future research directions.

\subsection{Comparison with Existing AD Development Approaches}

%\begin{table}[t]
%\centering
%\caption{Comparison of ADDT with Autonomous Driving Simulation Tools}
%\begin{tabular}{l|llll}
%\hline
%                            & Road Test & CarMaker~\cite{carmaker} & Carla~\cite{carla} & ADDT \\ \hline
%Hardware-in-Loop Simulation & yes       & no       & no    & yes  \\
%End-to-end Evaluation       & yes       & no       & yes   & yes  \\
%Fault Injection             & yes       & no       & no    & yes  \\
%Test Efficiency             & low       & high      & high  & high \\ \hline
%\end{tabular}
%\label{table:compare_platforms}
%\end{table}

\begin{table}[htbp]
\centering
\caption{Comparison of ADDT with Other Autonomous Driving Simulation Tools}
\newcolumntype{P}[1]{>{\centering\arraybackslash}p{#1}}
\renewcommand{\arraystretch}{1.2}
\scriptsize
\begin{tabular}{p{3.4cm}|P{1.0cm}P{1.0cm}P{1.2cm}P{1.2cm}P{1.2cm}P{1.2cm}}
\hline
\multirow{2.75}{*}{\textbf{Feature/Capability}} & \textbf{Physical Road Testing} & \textbf{\newline ADDT} & \textbf{NVIDIA DRIVE Sim~\cite{NVDriveSim}} & \textbf{Waymo\newline Simulation\newline City~\cite{WaymoSimCity}} & \textbf{IPG\newline Car-Maker~\cite{carmaker}} & \textbf{\newline Carla~\cite{carla}} \\
\hline
\textbf{Core Capabilities} & \multicolumn{6}{l}{} \\
%%\multicolumn{8}{l}{\textbf{Core Capabilities}} \\
Hardware-in-Loop Simulation & \cellcolor{green!100} & \cellcolor{green!25} & \cellcolor{green!25} & \cellcolor{red!100} & \cellcolor{green!100} & \cellcolor{red!100} \\
End-to-end Evaluation       & \cellcolor{green!100} & \cellcolor{green!25} & \cellcolor{green!25} & \cellcolor{green!25} & \cellcolor{green!25} & \cellcolor{green!25} \\
Fault Injection             & \cellcolor{red!100} & \cellcolor{green!100} & \cellcolor{red!100} & \cellcolor{red!100}  & \cellcolor{red!100} & \cellcolor{red!100} \\
Test Efficiency             & \cellcolor{red!100} & \cellcolor{green!25} & \cellcolor{green!25} & \cellcolor{green!100} & \cellcolor{green!25} & \cellcolor{green!25} \\
\hline
\textbf{Fidelity} & \multicolumn{6}{l}{} \\
%%\multicolumn{8}{l}{\textbf{Fidelity}} \\
Environment modeling        & \cellcolor{green!100} & \cellcolor{yellow!100} & \cellcolor{green!100} & \cellcolor{green!100} & \cellcolor{green!25} & \cellcolor{yellow!100} \\
Vehicle dynamics modeling   & \cellcolor{green!100} & \cellcolor{yellow!100} & \cellcolor{green!25} & \cellcolor{green!25} & \cellcolor{green!100} & \cellcolor{yellow!100} \\
Sensor physics modeling     & \cellcolor{green!100} & \cellcolor{yellow!100} & \cellcolor{green!100} & \cellcolor{green!100} & \cellcolor{green!25} & \cellcolor{yellow!100} \\
\hline
\textbf{Fault Testing} & \multicolumn{6}{l}{} \\
%%\multicolumn{8}{l}{\textbf{Fault Testing}} \\
Sensor fault injection      & \cellcolor{red!100} & \cellcolor{green!100} & \cellcolor{red!100} & \cellcolor{red!100} & \cellcolor{red!100} & \cellcolor{red!100} \\
Compute HW fault injection  & \cellcolor{red!100} & \cellcolor{green!100} & \cellcolor{red!100} & \cellcolor{red!100} & \cellcolor{red!100} & \cellcolor{red!100} \\
Latency variation analysis  & \cellcolor{yellow!100} & \cellcolor{green!100} & \cellcolor{green!25} & \cellcolor{yellow!100} & \cellcolor{green!25} & \cellcolor{yellow!100} \\
\hline
\textbf{Practical Considerations} & \multicolumn{6}{l}{} \\
%\multicolumn{8}{l}{\textbf{Practical Considerations}} \\
Cost/Open-source                        & \cellcolor{red!100} & \cellcolor{green!100} & \cellcolor{red!100} & \cellcolor{red!100} & \cellcolor{red!100} & \cellcolor{green!100} \\
\hline
\multicolumn{7}{l}{\textbf{Legend}} \\
\multicolumn{3}{l}{\cellcolor{green!100} \textbf{Excellent/Core strength}} & \multicolumn{4}{l}{\cellcolor{green!25} \textbf{Good}}\\
\multicolumn{3}{l}{\cellcolor{yellow!100} \textbf{Basic/Limited}} & \multicolumn{4}{l}{\cellcolor{red!100} \textbf{Unsafe or Not available}}\\
\hline
\end{tabular}
\label{table:addt_comparison}
\end{table}

Current autonomous driving (AD) development workflows, centered around physical road testing, log-replay analysis, and manual debugging, remain reactive, costly, and limited in scope. These approaches struggle to capture the diversity of real-world edge cases, particularly rare or emergent failures in complex and dynamic environments~\cite{liu2022rise, wu2023autonomy}. Physical testing is essential but resource-intensive, while log-replay systems~\cite{li2019aads} are constrained by historical data and cannot synthesize novel failure modes.

The ADDT framework surpasses these traditional tools in two critical aspects. First, its comprehensive digital twin infrastructure enables \textbf{proactive discovery of safety-critical faults}, such as sensor drift, compute errors, and latency spikes—before deployment. Second, its virtual-first methodology dramatically reduces reliance on physical testing, \textbf{delivering up to 90× efficiency improvement while supporting high-throughput, scalable validation}.

Specifically, ADDT integrates five core capabilities that distinguish it from existing simulators and toolchains: 

\begin{itemize}
    %\item \textbf{End-to-End Pipeline Evaluation:} Unlike tools such as CarMaker~\cite{carmaker}, which primarily targets planning and control subsystems, ADDT models the entire autonomy pipeline—including sensing, decision-making, actuation, and hardware infrastructure. This end-to-end visibility allows for tracing the cascading effects of faults (e.g., a sensor dropout leading to delayed control execution) across interconnected modules.

    \item \textbf{Integrated Fault Injection and Propagation Analysis:} ADDT includes built-in support for injecting repeatable, parameterized faults at both the sensor and compute levels. Competing tools like CARLA~\cite{carla} focus more on AD software but lack granular mechanisms for testing system resilience to hardware errors, such as memory corruption, CPU bit flips, or sensor misalignment.

    \item \textbf{Real-Time Latency Profiling:} In safety-critical systems, even small delays in perception or planning modules can lead to mission failure. ADDT’s built-in latency analysis tracks timing variations across system components under stress (e.g., dense traffic or environmental complexity), revealing conditions that push the system beyond its real-time thresholds—capabilities not provided by most simulators.

    \item \textbf{Hardware-in-the-Loop (HIL) Integration:} ADDT supports closed-loop HIL simulations, allowing developers to test real-time decision-making and compute resilience using actual hardware within virtual environments. This bridges the gap between simulation and deployment, enabling safer pre-deployment testing of control and compute units.

    \item \textbf{Scalability and Test Efficiency:} ADDT enables large-scale, high-throughput simulation across thousands of virtual scenarios—achieving up to 90× test efficiency improvement compared to physical road testing~\cite{yu2023autonomous}. It also eliminates logistical bottlenecks and safety risks associated with on-road prototyping.
\end{itemize}

Taken together, these capabilities position ADDT as a comprehensive solution for simulation-first, fault-aware AD system development. While existing tools offer valuable capabilities in isolation, they often lack the integration and fault-specific testing required to systematically address modern safety challenges. By providing a unified, extensible, and open-source platform, ADDT sets a new standard for proactive validation in autonomous driving system design.
A comparison between features of ADDT and existing toolchains  is illustrated in Table~\ref{table:addt_comparison}.

\subsection{Impact on Real-Time Safety and Reliability}

Latency variations, sensor faults, and hardware failures significantly threaten the safety and reliability of autonomous driving (AD) systems, particularly under demanding operational conditions. Our findings demonstrate that \textbf{latency spikes} frequently occur in scenarios involving dense object populations or highly dynamic environmental interactions, often exceeding the critical real-time threshold of 100 milliseconds. Such latency violations can severely compromise decision-making and control effectiveness, potentially leading to safety-critical incidents. To mitigate this issue, \textbf{the ADDT framework provides integrated real-time latency profiling across the entire AD pipeline}. By systematically identifying and evaluating latency-critical scenarios through high-fidelity simulations, ADDT enables developers to proactively optimize computational resources and timing constraints, ensuring adherence to real-time safety requirements.

Additionally, \textbf{sensor faults}—including frame drops and misalignment errors—tend to propagate through perception modules, substantially degrading environmental awareness and reducing mission success rates in complex and unpredictable driving conditions. These perception-level errors create significant challenges for downstream planning and control functions, impairing the vehicle's ability to accurately interpret and respond to rapidly evolving situations. \textbf{ADDT directly addresses this issue through built-in sensor fault injection mechanisms}, which allow precise simulation of realistic sensor errors. This capability helps developers understand fault propagation patterns, assess their impacts on overall mission performance, and develop robust perception and planning modules that can effectively handle sensor anomalies in real-world operations.

Furthermore, \textbf{compute-level faults}, particularly those originating within control units, disproportionately affect overall system performance compared to faults within perception or planning modules. This emphasizes the necessity of robust computational resilience, as failures at the control level directly compromise vehicle actuation and can lead to immediate and severe safety consequences. \textbf{The ADDT framework incorporates comprehensive compute-level fault injection and hardware-in-the-loop simulations}, enabling rigorous evaluation of computational reliability under controlled virtual conditions. By systematically stress-testing critical control modules, ADDT supports early identification of vulnerabilities and the development of fault-tolerant hardware-software designs that maintain robust performance even under adverse computational conditions.

Collectively, these targeted capabilities provided by ADDT \textbf{substantially enhance the proactive identification, evaluation, and mitigation of latency, sensor, and compute-related faults}. This systematic approach greatly improves the real-time safety and reliability of autonomous driving systems, reducing risks before deployment and strengthening overall system resilience.

\subsection{Limitations and Future Directions}

Although the ADDT framework significantly advances fault modeling and proactive validation of autonomous driving systems, certain limitations warrant further research. One key challenge involves accurately modeling fault propagation across interconnected modules, such as perception, planning, and control. The nonlinear and context-dependent nature of fault propagation complicates predicting downstream impacts, necessitating the development of advanced, nonlinear models to better represent these complex interactions and enhance overall system resilience.

Another critical area for improvement lies in scenario diversity and realism. While ADDT provides parameterizable scenarios, it may not yet achieve sufficient granularity to capture highly dynamic urban traffic conditions, including intricate pedestrian behaviors and unpredictable vehicle interactions. Future enhancements in scenario synthesis tools are essential for realistically replicating these complex urban environments, thus ensuring thorough validation under diverse real-world conditions.

Additionally, strengthening the integration of ADDT simulations with physical testing pipelines represents an important research direction. Combining comprehensive virtual validation with targeted real-world tests can validate simulation accuracy, further refine reliability assessments, and ultimately bridge the gap between simulation environments and actual deployment scenarios.

The establishment of standardized fault metrics is also crucial. Currently, universally accepted benchmarks and evaluation criteria for fault resilience remain underdeveloped. Efforts directed toward creating standardized metrics would enable consistent and fair comparisons of autonomous driving system performance, fostering industry-wide improvements.

Finally, extending ADDT to incorporate energy efficiency analysis presents significant potential for environmental benefits. While safety and fault resilience remain central objectives, integrating metrics for energy consumption would enable the optimization of computational resources, reducing energy usage throughout the development and operational phases of autonomous driving systems. This enhancement could substantially lower emissions and support the development of greener, more sustainable transportation solutions.

\subsection{Conclusion}
The ADDT framework represents a significant advancement over existing autonomous driving development methods and tools. Unlike traditional testing approaches, ADDT proactively detects critical corner-case failures before deployment through comprehensive fault modeling and high-fidelity simulations, offering substantial cost-efficiency and scalability. To further enhance ADDT, future research should focus on developing more precise fault propagation models, expanding the granularity and realism of urban scenario simulations, and integrating the framework with physical testing workflows. Beyond autonomous driving, ADDT’s simulation-first methodology holds considerable promise for advancing safety, reliability, and sustainability across the broader embodied AI industry, including robotics and intelligent automation.

\section{Materials and Methods}
\label{sec:mm}

The Autonomous Driving Digital Twin (ADDT) framework is designed as a high-fidelity closed-loop simulation platform to facilitate the design, validation, and optimization of autonomous driving (AD) systems. This section details the system architecture, fault injection methodologies, validation workflows, scenario design, and dataset integration. Key references have been included to enrich fault analysis depth, tool comparisons, and scenario-specific discussions.

\subsection{ADDT Architecture Design}
\begin{figure}[H]
    \centering
    \includegraphics[width=0.9\textwidth]{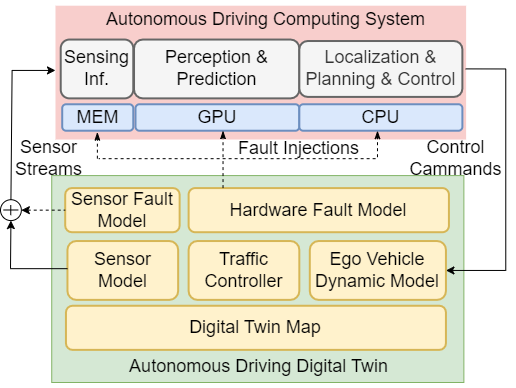}
    \caption{System architecture of the ADDT framework. The figure illustrates the key modules, including the digital twin map, traffic controller, sensor modeling, vehicle dynamics model, and fault injection mechanisms.} %These modules are integrated to enable high-fidelity closed-loop simulation with the AD computing system.}
    \label{fig:addt_arch}
\end{figure}

Fig.~\ref{fig:addt_arch} illustrates the ADDT system architecture. ADDT enables closed-loop simulation by providing interfaces to AD systems for raw sensor streams and control commands. Its autonomous driving simulator incorporates four core modules—a 3D digital twin map, sensor models, a vehicle dynamics model, and a traffic controller—to generate high-fidelity sensor data and simulate realistic traffic and vehicle behavior. ADDT also includes sensor and compute hardware fault injection modules for AD system safety assessment.

\subsubsection{Autonomous Driving Simulation}

\textbf{Digital twin map.}
The digital twin map use traditional traditional computer graphics (CG) models to manage the structural and physical characteristics of traffic infrastructure, such as roads, parking lots, and traffic lights. This enables traditional graphics pipeline to render sensor data. While traditional manual CG model creation is suitable for simple scenarios, it becomes costly for large-scale environments. To overcome this, we developed a method for constructing large-scale CG environments from autonomous vehicle high-definition (HD) maps~\cite{yu2022digitaltwin}. We extract geometric, structural, and type information from these HD maps, including road geometry, traffic facility types (e.g., road markers, traffic signs), and their positions. Then, we associate CG models with corresponding HD map objects based on extracted semantic information. Finally, we resize and place these CG models in the digital twin map using the extracted geometry and position data from the HD map.

\textbf{Traffic controller.}
The traffic controller manages traffic agents (e.g., vehicles and pedestrians) behavior for autonomous vehicle interaction. We utilize scenario description tools, such as OpenSCENARIO~\cite{openscenario}, to define agent movement and behavior within a scenario, enabling efficient test case composition and real-world scenario replication. 

\textbf{Vehicle dynamic and sensor simulation.}
For precise simulation of the ego vehicle's dynamic maneuvers, we leverage a commercial vehicle dynamic simulator~\cite{carmaker} within our ADDT framework, enabling co-simulation with other components. The ADDT system utilizes the game engine's image rendering pipeline to generate camera images. 

\subsubsection{Fault Modeling and Injection Mechanism.}
\begin{figure}[t]
\centering
  \includegraphics[trim=0 0 0 0, clip, width=1.\columnwidth]{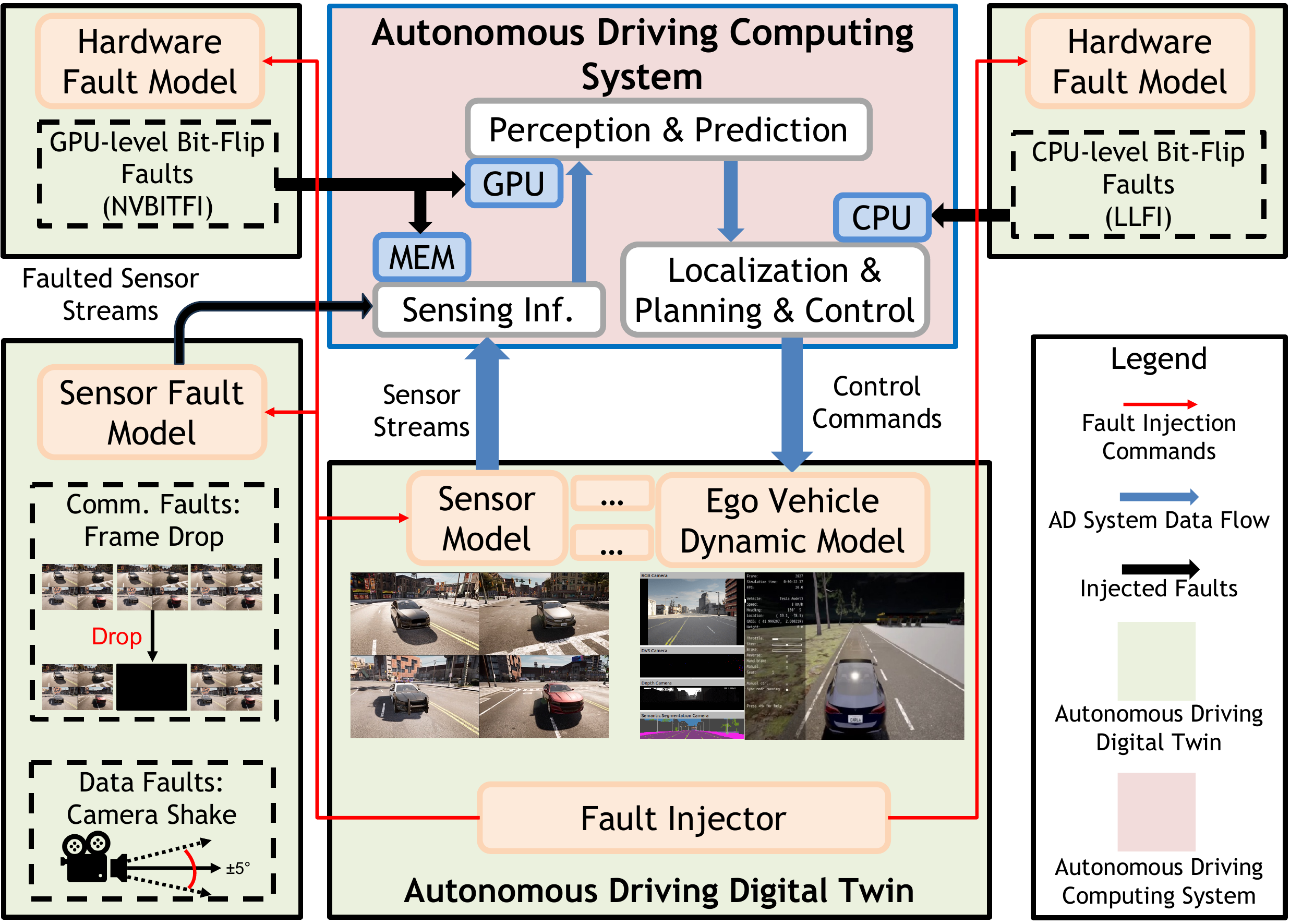}
\caption{Fault Injection Framework of the Proposed ADDT System.}
\vspace{8pt}
\label{fig:fault_injection}
\end{figure}

We identify three categories of faults in sensing and computing hardware, and integrate their injection mechanisms into our ADDT: sensor faults, CPU faults, and GPU faults. The proposed Fault Injection Framework is shown in Fig.~\ref{fig:fault_injection}.

\textbf{Sensor fault injection.} 
In an AD system, the sensing subsystem must provide well-calibrated multi-sensor data streams at constant frame rates. Sensor faults can degrade the quality of these data streams both temporally and spatially. To effectively simulate sensor faults, our ADDT system incorporates a sensor fault model that can inject faults into both domains.
\begin{enumerate}
    \item \textbf{Temporal domain error:} 
        The sensor fault model simulates temporal disruptions and frame drops within sensor data streams. We introduce controlled time delays to sensor streams by manipulating data timestamps. By imposing delays that exceed the perception model's out-of-order avoidance thresholds~\cite{FastBEV}, we effectively simulate frame drops.
       
    \item \textbf{Spatial domain error:}
        To simulate the spatial misalignment caused by real-world sensor position shifts, ADDT randomly introduces position and orientation shifts to ground truth extrinsic parameters. This enables precise control over misalignment magnitude for quantitative impact analysis on AD systems.
\end{enumerate}

\textbf{Computing hardware fault injection.} 
Computing hardware faults, such as memory and data path errors, pose significant safety risks to AD systems~\cite{wan2024vulnerability,jha2019ml}. To address these risks, the ADDT integrates a hardware fault co-simulator, enabling real-time, closed-loop simulation with fault injection. Given the heterogeneous nature of modern AD computing hardware, which includes CPUs and GPUs, our ADDT provides systematic fault injection mechanisms. For CPU fault injection, we utilize LLFI~\cite{lu2015llfi}, allowing us to simulate bit-flip faults at various points within the CPU's data path. For GPU fault injection, we employ NVBitFI~\cite{tsai2021nvbitfi}, a CUDA-level fault injection tool that simulates various faults on modern GPU architectures. 

To evaluate the impact of hardware bit-flip faults, our ADDT simulates fault injection at the register level of CPUs and GPUs.  
To ensure controllable fault injection, each fault in our ADDT is characterized by its location and injected value. During fault injection, the simulator modifies the variable values associated with the faulty registers, introducing errors into subsequent computations within the AD software. This fault injection method align with prior research~\cite{jha2019ml,wan2024vulnerability}. 

\subsection{Hardware-in-the-Loop (HIL) Setup}
\begin{figure}[t]
\centering
  \includegraphics[trim=0 0 0 0, clip, width=1.\columnwidth]{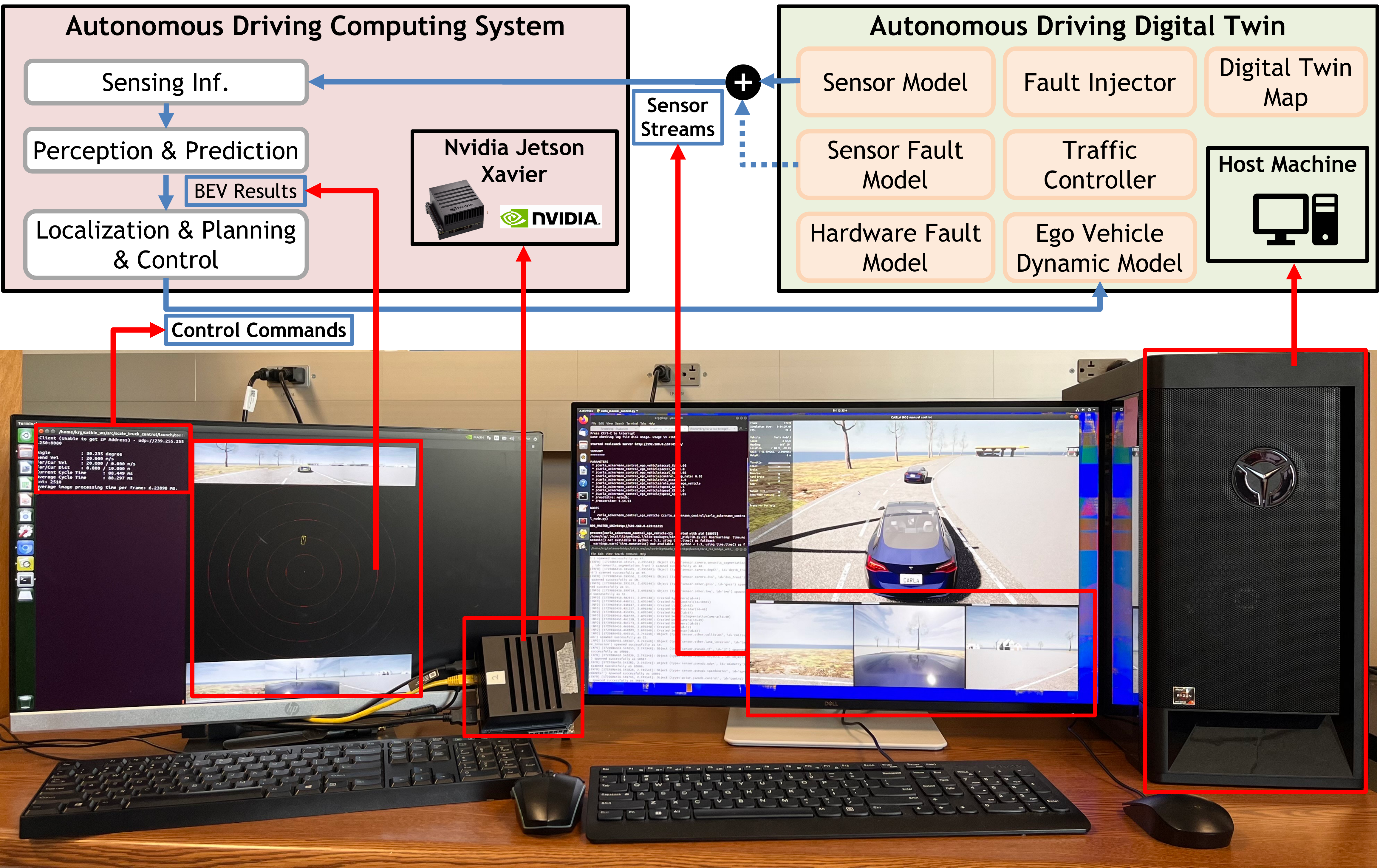}
\caption{ADDT Hardware-in-the-Loop(HIL) Setup}
\vspace{8pt}
\label{fig:ADDTframe}
\end{figure}

Fig.~\ref{fig:ADDTframe} illustrates the ADDT HIL platform, which comprises the ADDT simulation system and the AD computing system. In our setup, the two parts are connected via Ethernet. The ADDT sends sensor data to the AD system and receives control commands from the AD system to drive the vehicle within the simulator.

\subsubsection{ADDT simulation system.}
\textbf{Software.} The ADDT simulation is built upon Carla~\cite{carla}, an open-sourced AD simulator based on the Unreal Engine~\cite{ue}. %Methods from our prior work~\cite{yu2022digitaltwin} are integrated in the ADDT to enhance the realism of the digital twin map and sensor simulation. 
%A commodity vehicle dynamics simulator~\cite{carmaker} is integrated for precise vehicle dynamic simulation. 
Hardware fault modeling tools~\cite{lu2015llfi,tsai2021nvbitfi} are used for injecting hardware faults. OpenSCENARIO~\cite{openscenario} is used for specifying and parameterizing traffic scenarios.
\textbf{Hardware.} The ADDT software is deployed on a workstation with an Intel i7-9800X CPU and a NVIDIA GTX 2080 Ti GPU.

\subsubsection{AD computing system.}
\label{sec:ADCompSys}
\begin{figure}[H]
    \centering
    \includegraphics[width=1.0\textwidth]{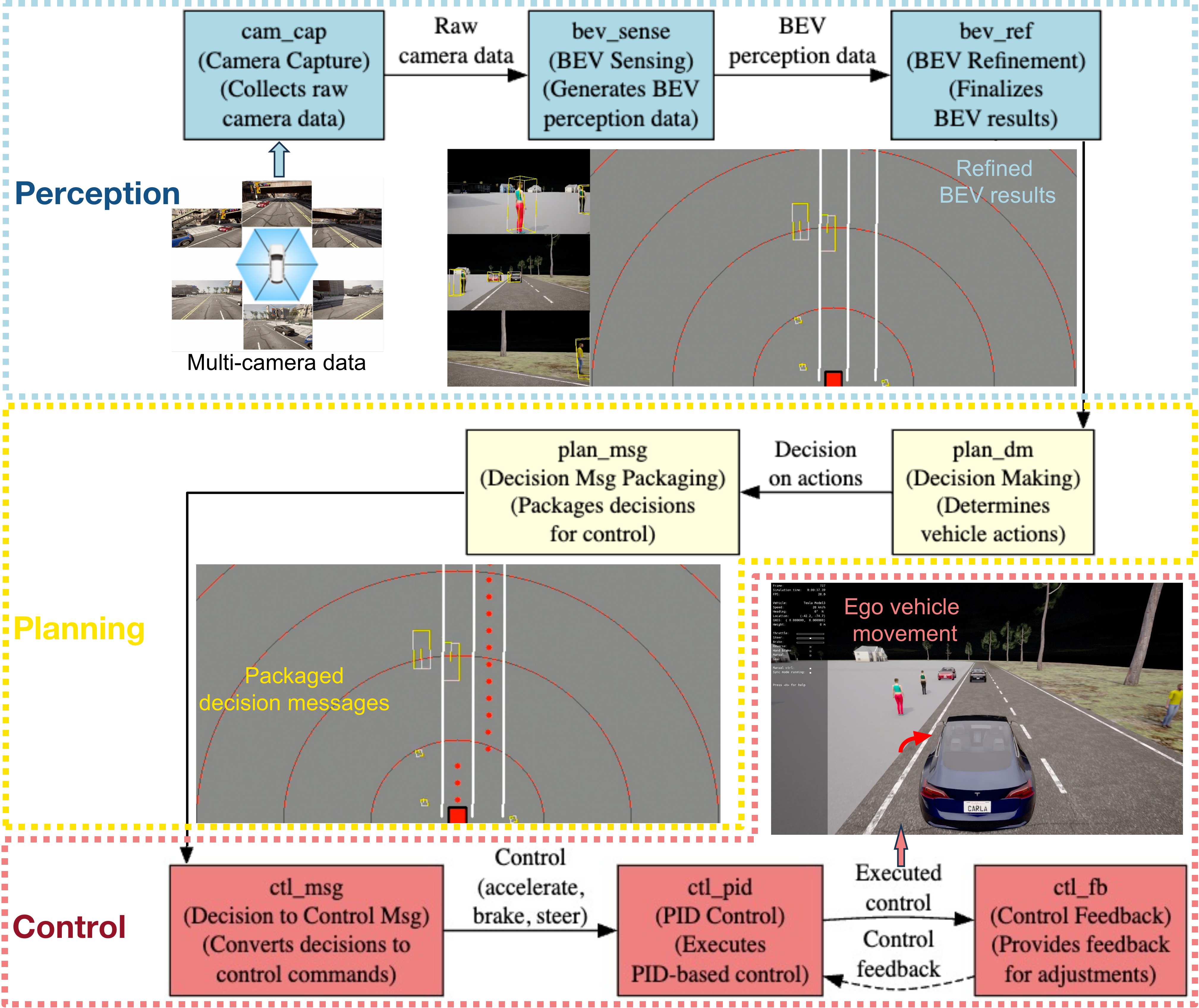}
    \caption{ROS node graph of the AD software.}
    \label{fig:graph_nodes}
\end{figure}

\textbf{Software.} 
Vision transformer-based algorithms and Bird's-Eye View (BEV) representations have become mainstream technologies in AD systems. To study the safety of modern AD systems, we developed an AD system based on BEV technology. Our AD system employ FastBEV~\cite{FastBEV} for environment perception, transforming objects from 2D image space into the BEV space. It uses a lattice-based method for path planning~\cite{bergman2020improved} and a PID-based method for vehicle control. Control commands are transmitted to the ADDT to control the ego-vehicle. ROS (Robot Operating System)~\cite{ROS} manages data communication between the components of the AD software and between the AD software and ADDT. Fig.~\ref{fig:graph_nodes} shows the ROS node graph of the AD software, illustrating the organization and interactions of ROS nodes within the AD software.
\textbf{Hardware.} The AD software is deployed on an Nvidia Jetson AGX Xavier. It is capable of processing six 480p images or three 720p images at a frame rate of 10 Hz and generating control commands at a frame rate of 100 Hz.

%\subsubsection{Driving Scenario Construction}
%Three primary categories of driving scenarios were synthesized using OpenSCENARIO~\cite{openscenario}:
%Three primary categories of driving scenarios were synthesized using OpenSCENARIO~\cite{openscenario}, leveraging its flexibility in defining complex traffic interactions and vehicle maneuvers. RoadRunner~\cite{RoadRunner} was utilized to design high-fidelity road networks, ensuring realistic lane structures, curvature, and traffic conditions. The synthesized scenarios include:

%\begin{itemize}
%    \item \textbf{Lane-Following with Emergency Braking:} Evaluated system latency and decision accuracy under sudden braking scenarios.
%    \item \textbf{High-Curvature Bends:} Simulated cornering maneuvers to assess sensor reliability and compute fault resilience.
%    \item \textbf{Lane-Changing and Overtaking:} Modeled dynamic traffic scenarios requiring rapid planning and precise control commands.
%\end{itemize}

\subsection{Data and Code Accessibility}
The ADDT codebase, encompassing the simulation software, AD software, fault injection tools, and HIL setup scripts, will be open-sourced under the AIRSHIP project (https://airs.cuhk.edu.cn/en/airstone) upon this paper's acceptance. Furthermore, the synthetic data used for evaluating latency variation, sensor faults, and compute faults will also be made available after acceptance.

%\begin{itemize}
%    \item \textbf{Code Repository:} \url{https://github.com/RichardYuanCR/AutoSim-Circuit}
%    \item \textbf{Datasets:} nuScenes, and OpenSCENARIO synthetic datasets.
%\end{itemize}

\subsection{Author Contributions}
First Author: Conceived the study, designed the digital twin framework, analyzed data, designed experiments, and wrote the manuscript. \\
Second Author: Implemented HIL and the fault injection mechanism, validated scenarios, performed statistical analysis, wrote the manuscript. \\
Third Author: Designed and developed the Fault Injection Mechanism, performed statistical analysis, and wrote the manuscript. \\
Fourth, Fifth, and Sixth Authors: Provided supervision, managed the project, acquired funding, and wrote and reviewed the manuscript. \\

\bibliographystyle{abbrv}
\footnotesize{%
\bibliography{ref}

\begin{thebibliography}{10}

\bibitem{accidents}
Autonomous vehicle accidents: Nhtsa crash data (2019-2024).
\newblock Accessed: 2025-03-15.

\bibitem{carmaker}
Carmaker.
\newblock \url{https://ipg-automotive.com/en/products-solutions/software/carmaker/}.
\newblock Accessed: 2023-12-10.

\bibitem{accidents2}
Drivers using tesla autopilot crashed hundreds of times, federal investigators say.
\newblock Accessed: 2025-03-15.

\bibitem{NVDriveSim}
Nvdrivesim.
\newblock \url{https://developer.nvidia.com/drive/simulation}.
\newblock Accessed: 2023-12-10.

\bibitem{openscenario}
Openscenario.
\newblock \url{https://www.asam.net/standards/detail/openscenario/v200/}.
\newblock Accessed: 2023-12-10.

\bibitem{ue}
Unreal engine.
\newblock \url{https://www.unrealengine.com/en-US/}.
\newblock Accessed: 2023-12-10.

\bibitem{WaymoSimCity}
Waymosimcity.
\newblock \url{https://waymo.com/blog/2021/07/simulation-city}.
\newblock Accessed: 2023-12-10.

\bibitem{aistrategy2023}
National ai r\&d strategic plan 2023.
\newblock \url{https://www.whitehouse.gov/wp-content/uploads/2023/05/National-Artificial-Intelligence-Research-and-Development-Strategic-Plan-2023-Update.pdf}, 2023.

\bibitem{bergman2020improved}
K.~Bergman, O.~Ljungqvist, and D.~Axehill.
\newblock Improved path planning by tightly combining lattice-based path planning and optimal control.
\newblock {\em IEEE Transactions on Intelligent Vehicles}, 6(1):57--66, 2020.

\bibitem{nuScenes}
H.~Caesar, V.~Bankiti, A.~H. Lang, S.~Vora, V.~E. Liong, Q.~Xu, A.~Krishnan, Y.~Pan, G.~Baldan, and O.~Beijbom.
\newblock nuscenes: A multimodal dataset for autonomous driving, 2020.

\bibitem{chen2025octocache}
P.~Chen, M.~Li, Z.~Wan, Y.-S. Hsiao, M.~Yu, V.~J. Reddi, and Z.~Liu.
\newblock Octocache: Caching voxels for accelerating 3d occupancy mapping in autonomous systems.
\newblock In {\em Proceedings of the 30th ACM International Conference on Architectural Support for Programming Languages and Operating Systems, Volume 2}, pages 704--718, 2025.

\bibitem{donner2009empirical}
C.~Donner, J.~Lawrence, R.~Ramamoorthi, T.~Hachisuka, H.~W. Jensen, and S.~Nayar.
\newblock An empirical bssrdf model.
\newblock In {\em ACM SIGGRAPH 2009 papers}, pages 1--10. 2009.

\bibitem{carla}
A.~Dosovitskiy, G.~Ros, F.~Codevilla, A.~Lopez, and V.~Koltun.
\newblock {CARLA}: {An} open urban driving simulator.
\newblock In {\em Proceedings of the 1st Annual Conference on Robot Learning}, pages 1--16, 2017.

\bibitem{gan2021eudoxus}
Y.~Gan, Y.~Bo, B.~Tian, L.~Xu, W.~Hu, S.~Liu, Q.~Liu, Y.~Zhang, J.~Tang, and Y.~Zhu.
\newblock Eudoxus: Characterizing and accelerating localization in autonomous machines industry track paper.
\newblock In {\em 2021 IEEE International Symposium on High-Performance Computer Architecture (HPCA)}, pages 827--840. IEEE, 2021.

\bibitem{gong2024lotus}
Y.~Gong, Y.~Wu, Z.~Zhan, P.~Zhao, L.~Liu, C.~Wu, X.~Tang, and Y.~Wang.
\newblock Lotus: learning-based online thermal and latency variation management for two-stage detectors on edge devices.
\newblock In {\em Proceedings of the 61st ACM/IEEE Design Automation Conference}, pages 1--6, 2024.

\bibitem{hsiao2023mavfi}
Y.-S. Hsiao, Z.~Wan, T.~Jia, R.~Ghosal, A.~Mahmoud, A.~Raychowdhury, D.~Brooks, G.-Y. Wei, and V.~J. Reddi.
\newblock Mavfi: An end-to-end fault analysis framework with anomaly detection and recovery for micro aerial vehicles.
\newblock In {\em 2023 Design, Automation \& Test in Europe Conference \& Exhibition (DATE)}, pages 1--6. IEEE, 2023.

\bibitem{hsiao2023silent}
Y.-S. Hsiao, Z.~Wan, T.~Jia, R.~Ghosal, A.~Mahmoud, A.~Raychowdhury, D.~Brooks, G.-Y. Wei, and V.~J. Reddi.
\newblock Silent data corruption in robot operating system: A case for end-to-end system-level fault analysis using autonomous uavs.
\newblock {\em IEEE Transactions on Computer-Aided Design of Integrated Circuits and Systems}, 43(4):1037--1050, 2023.

\bibitem{FastBEV}
B.~Huang, Y.~Li, E.~Xie, F.~Liang, L.~Wang, M.~Shen, F.~Liu, T.~Wang, P.~Luo, and J.~Shao.
\newblock Fast-bev: Towards real-time on-vehicle bird's-eye view perception, 2023.

\bibitem{jha2019ml}
S.~Jha, S.~Banerjee, T.~Tsai, S.~K. Hari, M.~B. Sullivan, Z.~T. Kalbarczyk, S.~W. Keckler, and R.~K. Iyer.
\newblock Ml-based fault injection for autonomous vehicles: A case for bayesian fault injection.
\newblock In {\em 2019 49th annual IEEE/IFIP international conference on dependable systems and networks (DSN)}, pages 112--124. IEEE, 2019.

\bibitem{AVFI}
S.~Jha, S.~S. Banerjee, J.~Cyriac, Z.~T. Kalbarczyk, and R.~K. Iyer.
\newblock Avfi: Fault injection for autonomous vehicles.
\newblock {\em 2018 48th Annual IEEE/IFIP International Conference on Dependable Systems and Networks Workshops (DSN-W)}, pages 55--56, 2018.

\bibitem{history}
T.~Krisher.
\newblock A brief history of eda.
\newblock \url{https://semiwiki.com/eda/1547-a-brief-history-of-eda/}, 2012.

\bibitem{li2019aads}
W.~Li, C.~Pan, R.~Zhang, J.~Ren, Y.~Ma, J.~Fang, F.~Yan, Q.~Geng, X.~Huang, H.~Gong, et~al.
\newblock Aads: Augmented autonomous driving simulation using data-driven algorithms.
\newblock {\em Science robotics}, 4(28):eaaw0863, 2019.

\bibitem{li2024bevformer}
Z.~Li, W.~Wang, H.~Li, E.~Xie, C.~Sima, T.~Lu, Q.~Yu, and J.~Dai.
\newblock Bevformer: learning bird's-eye-view representation from lidar-camera via spatiotemporal transformers.
\newblock {\em IEEE Transactions on Pattern Analysis and Machine Intelligence}, 2024.

\bibitem{liu2024intelligent}
S.~Liu.
\newblock {\em Intelligent Electric Vehicles}.
\newblock SAE International, 2024.

\bibitem{liu2022rise}
S.~Liu and J.-L. Gaudiot.
\newblock Rise of the autonomous machines.
\newblock {\em Computer}, 55(1):64--73, 2022.

\bibitem{liu2020creating}
S.~Liu, L.~Li, J.~Tang, S.~Wu, and J.-L. Gaudiot.
\newblock Creating autonomous vehicle systems.
\newblock {\em Synthesis Lectures on Computer Science}, 8(2):i--216, 2020.

\bibitem{liu2021brief}
S.~Liu, B.~Yu, Y.~Liu, K.~Zhang, Y.~Qiao, T.~Y. Li, J.~Tang, and Y.~Zhu.
\newblock Brief industry paper: The matter of time—a general and efficient system for precise sensor synchronization in robotic computing.
\newblock In {\em 2021 IEEE 27th Real-Time and Embedded Technology and Applications Symposium (RTAS)}, pages 413--416. IEEE, 2021.

\bibitem{PETR}
Y.~Liu, J.~Yan, F.~Jia, S.~Li, A.~Gao, T.~Wang, X.~Zhang, and J.~Sun.
\newblock Petrv2: A unified framework for 3d perception from multi-camera images, 2022.

\bibitem{lu2015llfi}
Q.~Lu, M.~Farahani, J.~Wei, A.~Thomas, and K.~Pattabiraman.
\newblock Llfi: An intermediate code-level fault injection tool for hardware faults.
\newblock In {\em 2015 IEEE International Conference on Software Quality, Reliability and Security}, pages 11--16. IEEE, 2015.

\bibitem{luo2019time}
Y.~Luo.
\newblock Time constraints and fault tolerance in autonomous driving systems.
\newblock {\em Tech. rep, Tech. Rep}, 2019.

\bibitem{ROS}
M.~Quigley, K.~Conley, B.~Gerkey, J.~Faust, T.~Foote, J.~Leibs, R.~Wheeler, and A.~Ng.
\newblock Ros: an open-source robot operating system.
\newblock volume~3, 01 2009.

\bibitem{seshia2022toward}
S.~A. Seshia, D.~Sadigh, and S.~S. Sastry.
\newblock Toward verified artificial intelligence.
\newblock {\em Communications of the ACM}, 65(7):46--55, 2022.

\bibitem{tsai2021nvbitfi}
T.~Tsai, S.~K.~S. Hari, M.~Sullivan, O.~Villa, and S.~W. Keckler.
\newblock Nvbitfi: Dynamic fault injection for gpus.
\newblock In {\em 2021 51st Annual IEEE/IFIP International Conference on Dependable Systems and Networks (DSN)}, pages 284--291. IEEE, 2021.

\bibitem{wan2023berry}
Z.~Wan, N.~Chandramoorthy, K.~Swaminathan, P.-Y. Chen, V.~J. Reddi, and A.~Raychowdhury.
\newblock Berry: Bit error robustness for energy-efficient reinforcement learning-based autonomous systems.
\newblock In {\em 2023 60th ACM/IEEE Design Automation Conference (DAC)}, pages 1--6. IEEE, 2023.

\bibitem{wan2025reca}
Z.~Wan, Y.~Du, M.~Ibrahim, J.~Qian, J.~Jabbour, Y.~Zhao, T.~Krishna, A.~Raychowdhury, and V.~J. Reddi.
\newblock Reca: Integrated acceleration for real-time and efficient cooperative embodied autonomous agents.
\newblock In {\em Proceedings of the 30th ACM International Conference on Architectural Support for Programming Languages and Operating Systems, Volume 2}, pages 982--997, 2025.

\bibitem{wan2024vulnerability}
Z.~Wan, Y.~Gan, B.~Yu, S.~Liu, A.~Raychowdhury, and Y.~Zhu.
\newblock The vulnerability-adaptive protection paradigm.
\newblock {\em Communications of the ACM}, 67(9):66--77, 2024.

\bibitem{wan2021survey}
Z.~Wan, B.~Yu, T.~Y. Li, J.~Tang, Y.~Zhu, Y.~Wang, A.~Raychowdhury, and S.~Liu.
\newblock A survey of fpga-based robotic computing.
\newblock {\em IEEE Circuits and Systems Magazine}, 21(2):48--74, 2021.

\bibitem{wang2025eai}
F.~Wang and S.~Liu.
\newblock Putting the smarts into robot bodies.
\newblock {\em Communications of the ACM}, 68(3), 2025.

\bibitem{wu2023autonomy}
S.~Wu, B.~Yu, S.~Liu, and Y.~Zhu.
\newblock Autonomy 2.0: The quest for economies of scale.
\newblock {\em Communications of the ACM}, 68(4), 2025.

\bibitem{yu2022digitaltwin}
B.~Yu, C.~Chen, J.~Tang, S.~Liu, and J.-L. Gaudiot.
\newblock Autonomous vehicles digital twin: A practical paradigm for autonomous driving system development.
\newblock {\em Computer}, 55(9):26--34, 2022.

\bibitem{yu2020building}
B.~Yu, W.~Hu, L.~Xu, J.~Tang, S.~Liu, and Y.~Zhu.
\newblock Building the computing system for autonomous micromobility vehicles: Design constraints and architectural optimizations.
\newblock In {\em 2020 53rd Annual IEEE/ACM International Symposium on Microarchitecture (MICRO)}, pages 1067--1081. IEEE, 2020.

\bibitem{yu2023autonomous}
B.~Yu, J.~Tang, and S.-S. Liu.
\newblock Autonomous driving digital twin empowered design automation: An industry perspective.
\newblock In {\em 2023 60th ACM/IEEE Design Automation Conference (DAC)}, pages 1--4. IEEE, 2023.

\end{thebibliography}
}
\end{document}